\documentclass[accepted]{uai2021}

\usepackage[T1]{fontenc}

\usepackage{algorithm}
\usepackage{adjustbox}
\usepackage[noend]{algpseudocode}
\usepackage{amsfonts}
\usepackage{amsmath}
\usepackage{amssymb}
\usepackage{amsthm}
\usepackage{comment}
\usepackage{cleveref}
\usepackage[mathscr]{eucal}
\usepackage[inline]{enumitem}
\usepackage{framed}
\usepackage{graphicx}
\usepackage{mathtools}
\usepackage{multirow}
\usepackage[authoryear,round]{natbib}
\usepackage{tabularx}
\usepackage[subrefformat=parens]{subcaption}
\usepackage{tikz}
\usepackage{thmtools}

\crefname{equation}{Eq.}{Eqs.}
\crefname{figure}{Fig.}{Figs.}
\crefname{table}{Table}{Tables}
\crefname{algorithm}{Algorithm}{Algorithms}
\crefname{section}{Sec.}{Secs.}

\usetikzlibrary{calc}
\usetikzlibrary{positioning}
\usetikzlibrary{shapes}

\usepackage{hyperref}
\hypersetup{
  hidelinks,
  colorlinks=true,
  linkcolor=black,
  citecolor=black,
}
\algnewcommand{\LineComment}[1]{\State \(\triangleright\) #1}

\declaretheorem[style=plain,numberwithin=section,qed=$\square$]{theorem}

\declaretheorem[style=definition,sibling=theorem]{definition}
\declaretheorem[style=definition,sibling=theorem]{example}
\declaretheorem[style=remark,sibling=theorem]{remark}

\allowdisplaybreaks

\DeclarePairedDelimiter{\abs}{\lvert}{\rvert}

\newcommand{\set}[1]{\{#1\}}
\newcommand{\defas}{\coloneqq}
\newcommand{\asdef}{\eqqcolon}
\newcommand{\Indicator}{\mathbf{1}}
\newcommand{\Indicate}[1]{\Indicator{[#1]}}
\newcommand{\Naturals}{\mathbb{N}}
\newcommand{\dist}[1]{\textrm{#1}}
\newcommand{\crp}{\dist{CRP}}
\newcommand{\domain}[1]{\textsc{#1}}
\newcommand{\hirm}{HIRM}

\title{Hierarchical Infinite Relational Model}

\author{Feras A.~Saad}
\author{Vikash K.~Mansinghka}
\affil{
  Massachusetts Institute of Technology \authorcr
  \normalfont Cambridge, MA, USA}

\begin{document}

\maketitle

\begin{abstract}
This paper describes the hierarchical infinite relational model
(HIRM), a new probabilistic generative model for noisy,
sparse, and heterogeneous relational data.
Given a set of relations defined over a collection of domains, the
model first infers multiple non-overlapping clusters of relations using a
top-level Chinese restaurant process.
Within each cluster of relations, a Dirichlet process mixture is then
used to partition the domain entities and model the probability
distribution of relation values.
The HIRM generalizes the standard infinite relational
model and can be used for a variety of data analysis tasks including
dependence detection, clustering, and density estimation.
We present new algorithms for fully Bayesian posterior inference via
Gibbs sampling.
We illustrate the efficacy of the method on a density estimation
benchmark of twenty object-attribute datasets with up to 18 million
cells and use it to discover relational structure in
real-world datasets from politics and genomics.
\end{abstract}


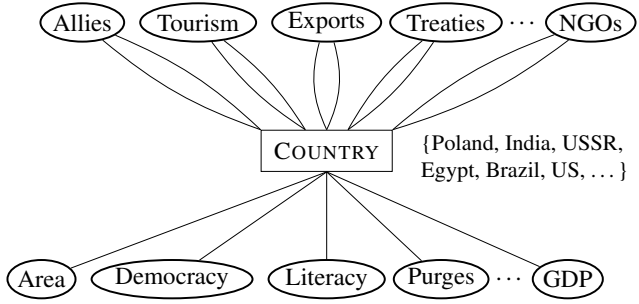
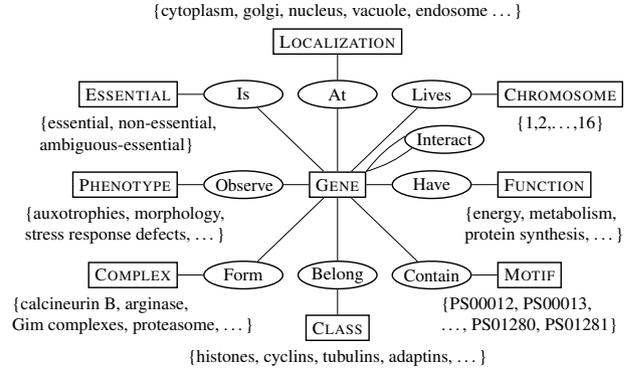
\begin{figure*}
\centering
\begin{subfigure}[b]{.495\linewidth}
\centering
\begin{adjustbox}{max width=\linewidth}
\begin{tikzpicture}
\def\dist{1.25}

\tikzset{dom/.style = {rectangle,inner sep=2pt,draw=none}}
\tikzset{relation/.style = {ellipse,thick,draw=black,minimum height=.55cm,inner sep=0pt,align=center}}

\node[name=country,dom,inner sep=5pt, draw=black,
  label={[label distance=2,anchor=west,yshift=-.1cm]right:{\footnotesize\begin{tabular}{l} \{Poland, India, USSR, \\ Egypt, Brazil, US, \dots\} \end{tabular}}},
  ]{{\domain{Country}}};

\node[name=literacy,relation,below =\dist of country,xshift=0cm]{Literacy};
\node[name=democracy,relation,left=.2 of literacy]{Democracy};
\node[name=area,relation,left=.2 of democracy]{Area};
\node[name=purges,relation,right=.1 of literacy.east,anchor=west]{Purges};
\node[name=gdp,relation,right=.6 of purges]{GDP};
\node[name=dots,anchor=center,align=center,inner sep=0pt] at ($(purges.east)!0.5!(gdp.west)$) {\dots};

\draw (country.south) -- (literacy);
\draw (country.south) -- (democracy);
\draw (country.south) -- (area);
\draw (country.south) -- (purges);
\draw (country.south) -- (gdp);

\node[name=rliteracy,relation,above=\dist of country,xshift=0cm]{Exports};
\node[name=rdemocracy,relation,left=.2 of rliteracy]{Tourism};
\node[name=rarea,relation,left=.2 of rdemocracy]{Allies};
\node[name=rpurges,relation,right=.1 of rliteracy.east,anchor=west]{Treaties};
\node[name=rgdp,relation,right=.6 of rpurges]{NGOs};
\node[name=rdots,anchor=center,align=center,inner sep=0pt] at ($(rpurges.east)!0.5!(rgdp.west)$) {\dots};


\draw (rarea) edge [bend left=10] (country.north west);
\draw (rarea) edge [bend right=10] (country.north west);

\draw (rdemocracy) edge [bend left=10] (country.135);
\draw (rdemocracy) edge [bend right=10] (country.135);

\draw (rliteracy) edge [bend left=20] (country.north);
\draw (rliteracy) edge [bend right=20] (country.north);

\draw (rpurges) edge [bend left=10] (country.45);
\draw (rpurges) edge [bend right=10] (country.45);

\draw (rgdp) edge [bend left=10] (country.north east);
\draw (rgdp) edge [bend right=10] (country.north east);

\end{tikzpicture}
\end{adjustbox}
\captionsetup{aboveskip=20pt}
\caption{Relational system for political data~\citep{rummel1999}}
\label{fig:system-nations}
\end{subfigure}\hfill\vline%
\begin{subfigure}[b]{.495\linewidth}
\centering
\begin{adjustbox}{max width=\linewidth}
\begin{tikzpicture}
\def\dist{1.25}
\def\ldist{0pt}

\tikzset{dom/.style = {rectangle,thick,minimum height=.5cm,draw=black}}
\tikzset{attribute/.style = {draw=black,thick,minimum height=.55cm,anchor=south}}
\tikzset{relation/.style = {ellipse,thick,draw=black,minimum height=.55cm,inner sep=0pt,text width=1.1cm,align=center}}

\node[name=gene,dom]{{\domain{Gene}}};

\node[name=belong,relation,below=\dist of gene]{Belong};
\node[name=class,dom,below=.5 of belong,
  label={[label distance=-\ldist]below:{\begin{tabular}{l} \{histones, cyclins, tubulins, adaptins, \dots\} \end{tabular}}},
  ]{{\domain{Class}}};

\node[name=have,relation,right=.5 of gene]{Have};
\node[name=function,dom,right=.5 of have,
  label={[label distance=-\ldist]below:{\begin{tabular}{l} \{energy, metabolism, \\ protein synthesis, \dots \} \end{tabular}}},
  ]{{\domain{Function}}};

\node[name=contain,relation] at (belong -| have) {Contain};
\node[name=motif,dom,right=.5 of contain,
  label={[label distance=-\ldist]below:{\begin{tabular}{l} \{PS00012, PS00013, \\  \dots, PS01280, PS01281\} \end{tabular}}},
  ]{{\domain{Motif}}};

\node[name=observe,relation,left=.5 of gene]{Observe};
\node[name=phenotype,dom,left=.5 of observe,
  label={[label distance=-\ldist]below:{\begin{tabular}{l} \{auxotrophies, morphology, \\ stress response defects, \dots \} \end{tabular}}},
  ]{{\domain{Phenotype}}};

\node[name=form,relation] at (belong -| observe) {Form};
\node[name=complex,dom,left=.5 of form,
    label={[label distance=-\ldist]below:{\begin{tabular}{l} \{calcineurin B, arginase, \\ Gim complexes, proteasome, \dots \} \end{tabular}}},
  ]{{\domain{Complex}}};

\node[name=at,relation,above=\dist of gene]{At};
\node[name=localization,dom,above=.5 of at,
  label={[label distance=-\ldist]above:{\begin{tabular}{l} \{cytoplasm, golgi, nucleus, vacuole, endosome \dots \} \end{tabular}}},
  ]{{\domain{Localization}}};

\node[name=is,relation] at (at -| observe) {Is};
\node[name=essential,dom,left=.5 of is,
  label={[label distance=-\ldist]below:{\begin{tabular}{l} \{essential, non-essential, \\ ambiguous-essential\} \end{tabular}}},
  ]{{\domain{Essential}}};

\node[name=lives,relation] at (is -| have) {Lives};
\node[name=chromosome,dom,right=.5 of lives,
  label={[label distance=-\ldist]below:{\begin{tabular}{l} \{1,2,\dots,16\} \end{tabular}}},
  ]{{\domain{Chromosome}}};

\node[name=interact,relation,xshift=.25cm] at ($(have)!0.5!(lives)$) {Interact};

\draw (gene) -- (have) -- (function);
\draw (gene) -- (belong) -- (class);
\draw (gene) -- (contain) -- (motif);
\draw (gene) -- (observe) -- (phenotype);
\draw (gene) -- (form) -- (complex);
\draw (gene) -- (is) -- (essential);
\draw (gene) -- (at) -- (localization);
\draw (gene) -- (lives) -- (chromosome);

\draw (gene.north east) edge [bend left=10] (interact.175);
\draw (gene.north east) edge [bend right=10] (interact.195);

\end{tikzpicture}
\end{adjustbox}
\caption{Relational system for genomics data~\citep{cheng2001}}
\label{fig:system-genes}
\end{subfigure}
\caption{
Two relational systems that we analyze using the \hirm{} in this paper.
Domains are in boxes, relations in ellipses, and domain entities
between curly braces.
\subref{fig:system-nations} One domain, five unary relations,
and five binary relations.
\subref{fig:system-genes} Nine domains and nine binary relations.
Unary relations represent ``attributes'' while binary relations and higher represent ``interactions''.
}
\label{fig:systems}
\end{figure*}

\section{Introduction}

Learning models for relational data is a widely
studied problem that arises in a number of settings such as business
intelligence~\citep{chaudhuri2011}, social networks~\citep{carrington2005},
bioinformatics~\citep{rual2005}, and recommendation systems~\citep{su2009},
amongst many others~\citep{dzeroski2001}.
In this setting, we observe attributes and interactions among a set
of entities and our goal is to learn models that are useful for
explaining or making predictions about the entities, their attributes,
and/or their interactions.
\Cref{fig:systems} shows two examples of relational systems
for political and genomics data.
For politics (\cref{fig:system-nations}), one problem could be to discover
what attributes of a particular country and interactions with other
countries are likely to make it an attractive tourist hub.
In genomics (\cref{fig:system-genes}), our goal might be to predict what
complexes a particular gene is likely to form, given information
about its motifs, functions, and interactions with other genes.
This paper addresses the problem of automatically learning
probabilistic models for a variety of relational systems
given a dataset of noisy and possibly sparse observations.

Learning probabilistic structure is an exceptionally
difficult task~\citep{daly2011}.
One approach to simplifying the learning problem is to posit a
collection of hidden variables that both explain and decouple
the relationships between observed variables.
Using Bayesian nonparametrics, both the values and dimensionality of
these hidden variables can be automatically inferred from data.
This approach is commonly used for modeling relational
data~\citep{kemp2006,xu2006,roy2008,sutskever2009,kim2013,nakano2014,xuan2017,fan2018}:
refer to~\citet{fan2020} for a recent survey on developments in the field.
Our paper builds on the infinite relational
model~\citep[IRM;][]{kemp2006,xu2006}, a widely used and flexible
Bayesian nonparametric method that applies to a variety relational
systems.
The IRM is a cluster-based model: informally, to decide whether a binary
relation $R$ holds between a pair of entities $i$ and $j$, the IRM
flips a coin whose weight depends on the (latent)
cluster assignments of $i$ and $j$.
A strength of the IRM, which we review in \cref{sec:irm}, is its
ability to extract meaningful partitions from observational data.
However, as we identify in \cref{sec:limitations}, two
limitations inherent to the IRM's inductive bias make the model
\begin{enumerate*}[label=(\roman*)]
\item susceptible to combinatorial over-clustering; and

\item fail to discover certain predictive structure between dependent but
non-identically distributed relations,
\end{enumerate*}
which can both result in an inaccurate overall model of the data.

To address these limitations, this paper introduces  the hierarchical
infinite relational model (\hirm{}) in \cref{sec:hirm}, a new method that combines the
flexibility of the IRM with a structure learning prior that infers
subsets of relations that are probably independent of one another.
By allowing different relations to be explained by different
partitions, the \hirm{} specifies a large hypothesis space that
includes the standard IRM in addition to compact models of the data
that can only be approximated by an IRM using a combinatorially large
number of clusters.
The evaluations in \cref{sec:evaluations} show that the \hirm{}
makes more accurate predictions and discovers more fine-grained
clustering structure as compared to the IRM, while retaining a
flexible framework for automatic Bayesian structure discovery in a
variety of relational systems.

\section{Infinite Relational Model}
\label{sec:irm}

We begin with a review of the IRM, using a slightly more general
definition of ``relations'' than was originally described
in~\citet{kemp2006} or~\cite{xu2006}.

\begin{definition}
\label{def:system}
A \textit{relational system} $S$ consists of $n$ domains $D_1, \dots, D_n$
and $m$ relations $R_1, \dots, R_m$.
Each \textit{domain} $D_i$ ($1 \le i \le n$) is a countably
infinite set of distinct \textit{entities} $\set{e^i_{1}, e^i_{2}, \dots}$.
Each \textit{relation} $R_k$ ($1 \le k \le m$) is a map
from the Cartesian product of $t_k$ domains to an arbitrary codomain $C_k$.
The symbol $d_{ki}$ ($1 \le k \le m$, $1 \le i \le t_k$) denotes the domain index of
the $i$-th argument of $R_k$.
\end{definition}

\begin{example}
Suppose system $S$ has $n = 4$ domains $D_1$, $D_2$, $D_3$, $D_4$, and
$m=3$ relations $R_1, R_2, R_3$; with
\begin{align*}
  R_1 &: D_1 \times D_1 \to \set{0,1}, \span\span\\
  R_2 &: D_1 \times D_3 \times D_4 \to \set{1,2,\dots}, \span\span\\
  R_3 &: D_2 \to (-\infty, \infty). \span\span\\
\shortintertext{In this system, we have}
t_1 = 2; &\quad d_{11} = 1, d_{12} = 1; &&C_1 = \set{0,1}; \notag \\
t_2 = 3; &\quad d_{21} = 1, d_{22} = 3, d_{23} = 4; &&C_2 = \set{1,2,\dots}; \notag\\
t_3 = 1; &\quad d_{31} = 2; &&C_3 = (-\infty, \infty). \notag
\end{align*}
$R_1$ is a binary relation taking binary values, $R_2$ is a ternary
relation taking positive integer values, and $R_3$ is a unary relation
taking real values.
\end{example}

\begin{remark}
\label{remark:relation-index}
To simplify notation, for a given relation
$R: D_1 \times \dots \times D_n \to C$ and entity indexes
$i_1,\dots,i_n \in \Naturals$,
we will write $R(i_1, \dots, i_n)$ to mean
$R(e^1_{i_1}, \dots, e^n_{i_n})$.
\end{remark}


\begin{figure*}[t]
\setlength{\FrameSep}{0pt}
\tikzset{pic/.style={inner sep=0pt, draw=none, thick}}
\begin{subfigure}[b]{.3333\textwidth}
\begin{framed}
\begin{tikzpicture}
\node[name=f1,pic]{\includegraphics[width=\linewidth]{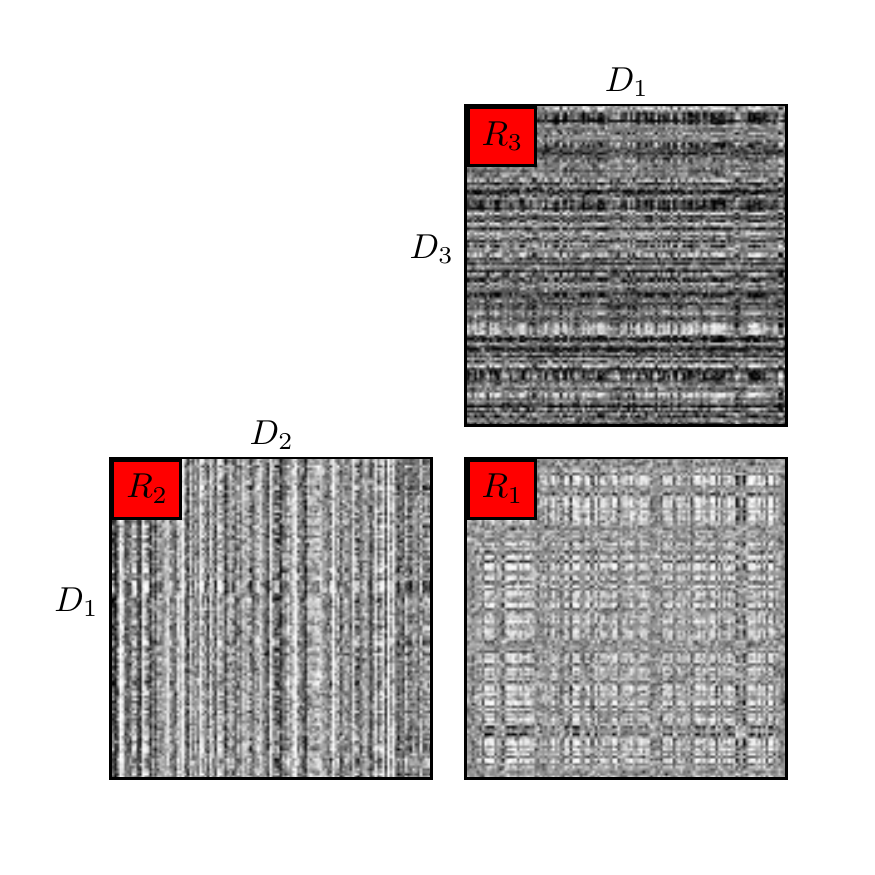}};%
\node[name=f2,
  align=center,
  draw=none,
  below left = 1.2 and 1.5 of f1.north,
  anchor=center]{\scriptsize
  $\begin{aligned}
  R_1&:D_1{\times}D_1\to\set{0,1} \\[-3pt]
  R_2&:D_1{\times}D_2\to\set{0,1} \\[-3pt]
  R_3&:D_3{\times}D_1\to\set{0,1}
  \end{aligned}$
};
\node[name=f3,align=center,
  below left = .5 and 1.25 of f1.north,
  anchor=center
] {\scriptsize\bfseries Relational System};
\end{tikzpicture}
\end{framed}
\captionsetup{skip=-5pt}
\caption{Relational System and Dataset}
\label{fig:hirm-samples-data}
\end{subfigure}%
\begin{subfigure}[b]{.3333\textwidth}
\begin{framed}
\begin{tikzpicture}
\node[name=f1,pic]{\includegraphics[width=\linewidth]{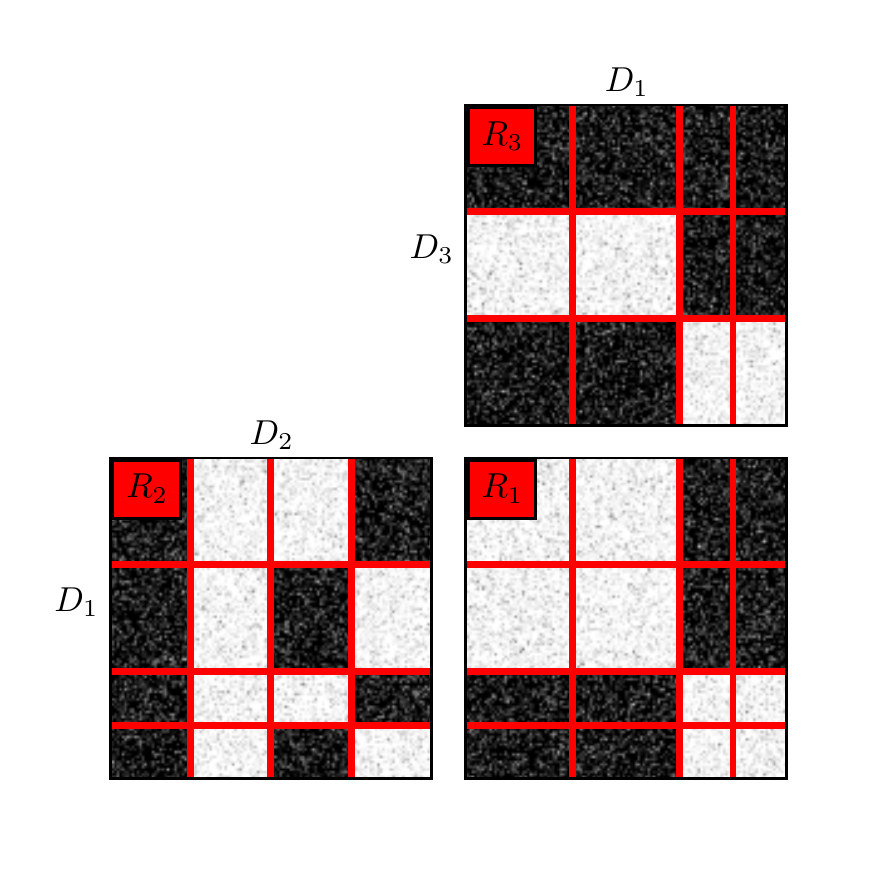}};%
\node[name=f2,
  align=center,
  circle,
  draw=black,
  label={left:$R_1$},
  label={below:$R_3$},
  label={right:$R_2$},
  below left = 1 and 1.4 of f1.north,
  anchor=center]{};
\node[name=f3,align=center,
  below left = .5 and 1.25 of f1.north,
  anchor=center
] {\scriptsize\bfseries Relation Partition};
\end{tikzpicture}
\end{framed}
\captionsetup{skip=-5pt}
\caption{IRM (No Structural Independences)}
\label{fig:hirm-samples-1}
\end{subfigure}%
\begin{subfigure}[b]{.3333\textwidth}
\begin{framed}
\begin{tikzpicture}
\node[name=f1,pic]{\includegraphics[width=\linewidth]{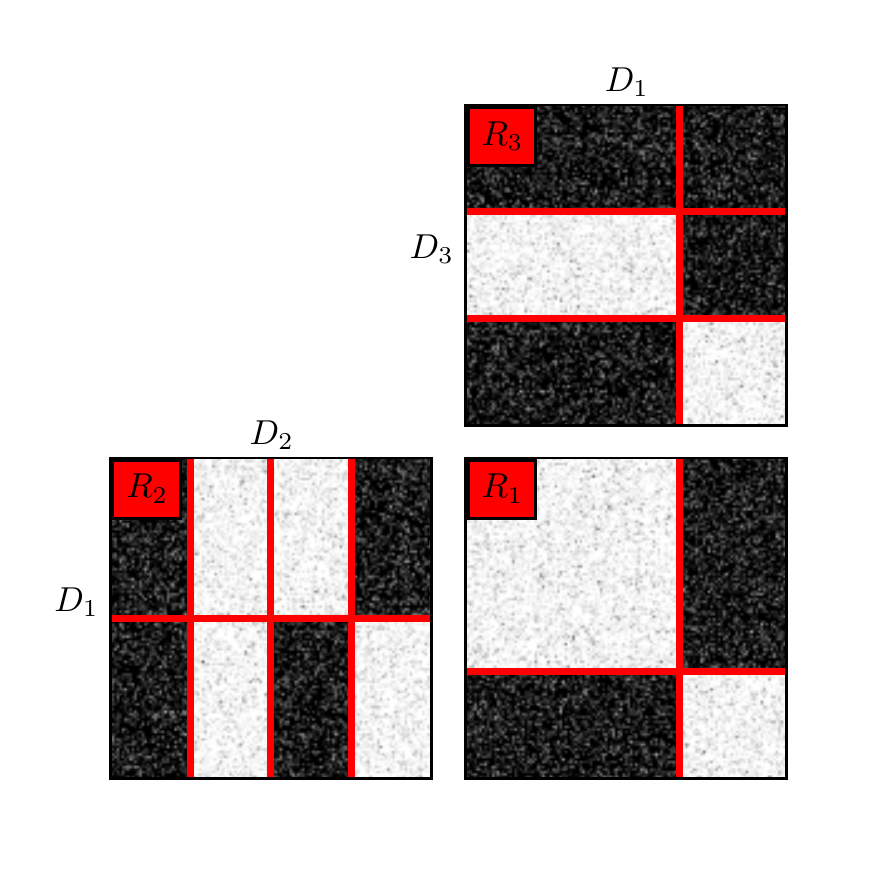}};%
\node[name=f2,
  align=center,
  circle,
  draw=black,
  label={left:$R_1$},
  label={below:$R_3$},
  below left = 1 and 1.85 of f1.north,
  anchor=center]{};
\node[name=f4,
  align=center,
  circle,
  draw=black,
  label={right:$R_2$},
  right = .85 of f2.center,
  anchor=center]{};
\node[name=f3,align=center,
  below left = .5 and 1.25 of f1.north,
  anchor=center
] {\scriptsize\bfseries Relation Partition};
\end{tikzpicture}
\end{framed}
\captionsetup{skip=-5pt}
\subcaption{\hirm{} ($(R_1,R_3)\perp R_2$)}
\label{fig:hirm-samples-2}
\end{subfigure}%
\captionsetup{skip=5pt}
\caption{Comparing posterior inferences from the standard IRM~\citep{kemp2006} and HIRM.
\subref{fig:hirm-samples-data} Three relation signatures and
an observed dataset.
\subref{fig:hirm-samples-1} The IRM forces all relations to be
dependent and learns, for each domain, a single clustering
(vertical/horizontal red lines) that is identical across all the
relations.
This conservative assumption by the IRM leads to an over-clustering of
the domain $D_1$ that participates in all three relations,
which in turn creates combinatorially many spurious clusters
than are required to explain the data.
\subref{fig:hirm-samples-2} The \hirm{} finds sufficient evidence
for the probable independence of $R_2$ with $(R_1, R_3)$ and learns
separate a domain clustering for $D_1$ within each of the two blocks of
the relation partition, leading to a more succinct explanation of the data.
While the clusterings in both~\subref{fig:hirm-samples-1} and~\subref{fig:hirm-samples-2}
are in the \hirm{} hypothesis space, the clustering
in~\subref{fig:hirm-samples-2} is ${\approx}\,7{\times}{10}^{13}$ times more likely
under the posterior, indicating that the independence structure and
domain clusterings inferred by the \hirm{} is substantially more likely
than the full dependence structure imposed by the IRM
in~\subref{fig:hirm-samples-1}.}
\label{fig:hirm-samples}
\end{figure*}
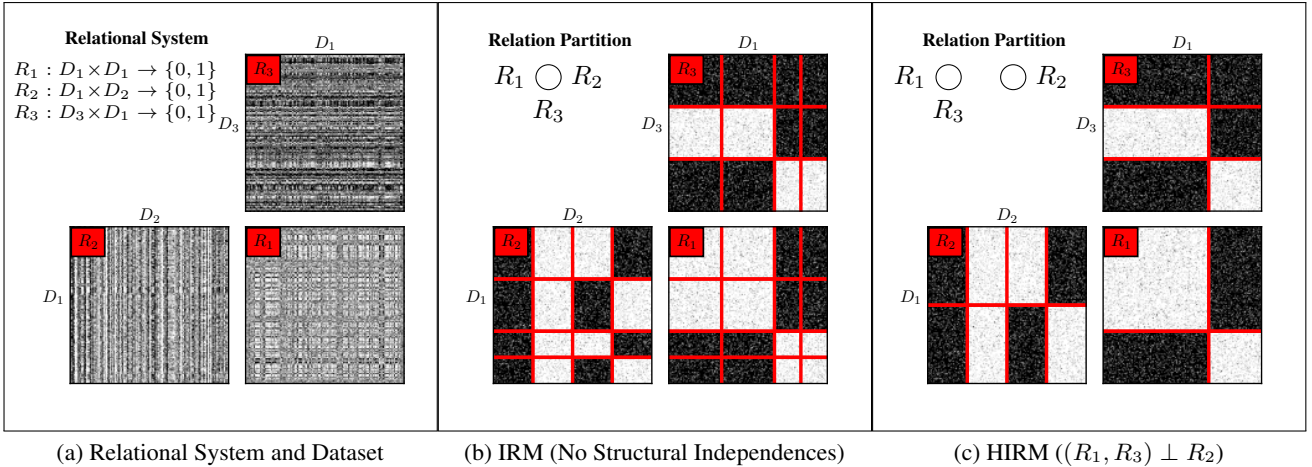

Consider a system $S$ with $n$ domains and $m$ relations.
For each $i = 1,\dots,n$, the IRM assumes that entities
$\set{e^i_1, e^i_2, \dots}$ in domain $D_i$
are associated with integer cluster assignments
$\set{z^i_1, z^i_2, \dots} \asdef z^i$.
The IRM defines a joint probability distribution over cluster
assignments and relation values
with the following factorization structure:
\begin{align}
\begin{aligned}[t]
&P(z^1, \dots, z^n, R_1, \dots, R_m) \\
&\qquad = \prod_{i=1}^{n}P(z^i)\prod_{k=1}^{m}P(R_k \mid z^1, \dots, z^n).
\end{aligned}
\label{eq:irm-factorization}
\end{align}
To allow the IRM to discover an arbitrary number of clusters for
each domain $D_i$, the cluster assignments $z^i$
for the entities are given a nonparametric prior that
assigns a positive probability to all possible partitions using the
Chinese restaurant process \citep[CRP;][]{aldous1985}.
For each $i=1,\dots,n$, the cluster assignment probabilities $P(z^i) = P(z^i_1, z^i_2, \dots)$
in \cref{eq:irm-factorization}
are defined inductively with $z^i_1 \defas 1$, and for $l \ge 2$
\begin{align}
P(z^i_l = j \mid z^i_1, \dots, z^i_{l-1}) \propto
  \begin{cases}
    {n_j}    & \mbox{if } 1 \le j \le M \\
    {\gamma} & \mbox{if } j = M + 1,
  \end{cases}
\label{eq:crp-probs}
\end{align}
where
  $n_j \defas \sum_{c=1}^{l-1}\Indicate{z^i_c = j}$ is the number
  of previous entities at cluster $j$;
  $M \defas \max\set{z^i_1,\dots,z^i_{l-1}}$ is the
    number of clusters among the first $l-1$ entities; and
  $\gamma > 0$ is a concentration parameter.
The cluster assignment vectors $z^1, \dots, z^n$ across the $n$ domains are mutually
independent, each drawn from a CRP (\cref{eq:crp-probs}).
Next, for each relation $R_k$ $(1 \le k \le m)$, a set of parameters
$\theta_k(j_1,\dots,j_{t_k})$ is used to dictate the distribution of
$R_k(i_1,\dots,i_{t_k})$, where $j_1, \dots, j_{t_k}, i_1, \dots,
i_{t_k} \in \Naturals$.
The value of a relation depends only the cluster assignments,
i.e., $R_k(i_1,\dots,i_{t_k})$ and $R_k(i'_1,\dots,i'_{t_k})$
share the same parameter whenever
$z^{d_{kl}}_{i_l} = z^{d_{kl}}_{i'_l}$
for each $l = 1, \dots,t_k$.
Thus, for domain index $i=1,\dots,n$;
  relation index $k=1,\dots,m$;
  entity indexes $i_1, \dots, i_{t_k} \in \Naturals$; and
  cluster indexes $j_1, \dots, j_{t_k} \in \Naturals$,
  the generative model of the IRM is given by:
\begin{align}
\set{z^i_1,z^i_2,\dots} &\sim \crp(\gamma_i)
  \label{eq:irm-crp} \\
\theta_k(j_1,\dots,j_{t_k}) &\sim \pi_k(\lambda_k)
  \label{eq:irm-param} \\
R_k(i_1,\dots,i_{t_k}) &\sim
  L_k(\theta_k({z^{d_{k1}}_{i_1}, \dots, z^{d_{kt_k}}_{i_{t_k}}})),
  \label{eq:irm-relation} &
\end{align}
where $(\set{\gamma_i}_{i=1}^{n}, \set{\lambda_k}_{k=1}^{m})$ are model
hyperparameters.
\Cref{eq:irm-relation} ensures items within a cluster
are generated by the same parameter.
%
%
The prior $\pi_k$ and likelihood $L_k$ distributions in
\cref{eq:irm-param,eq:irm-relation}
can be set
depending on the codomain $C_k$ of $R_k$ (e.g., beta-Bernoulli for
binary data, gamma-Poisson for counts, chisquare-normal for real
values, etc.).
\citet{kemp2006} used the IRM to discover structure
in a variety of real-world relational systems that appear quite different on the
surface, including:
\begin{enumerate}[label=(\alph*)]
\item Random graphs, with
  one domain $D$ for the vertices and one relation $R: D \times D \to
  \set{0,1}$ for the edges.

\item Object-attribute data, with one relation
  $R: D_1 \times D_2 \to \set{0,1}$, where $R(i,j)=1$ iff item
  $e^1_i$ has attribute $e^2_j$.

\item Systems with multiple attributes and interactions, where, for example,
  $D_1$ are countries,
  $D_2$ are attributes; and
  $D_3$ are interactions; so that
  $R_1: D_1 \times D_2 \to \set{0,1}$ models attributes and
  $R_2: D_1 \times D_1 \times D_3 \to \set{0,1}$ models interactions,
  where $R_2(i,j,k)\,{=}\,1$ iff
  countries $e^1_i$ and $e^1_j$ perform interaction $e^3_k$.
\end{enumerate}

\newpage

\section{Limitations of the IRM}
\label{sec:limitations}

We next describe two limitations in the standard IRM that arise when
using the model in practice, motivating the hierarchical structure
learning prior that we introduce in \cref{sec:hirm}.

\subsection{Enforcing Shared Domain Clusterings Leads to Overfitting}
\label{sec:limitations-overfit}

A key assumption of the IRM is that each domain $D_i$
has a single clustering $z^i = \set{z^i_1,z^i_2,\dots}$ that globally
dictates the partition of its entities $\set{e^i_1, e^i_2, \dots}$.
The same cluster assignments $z^i$ are used for all of
relations $R_1, \dots, R_m$ in which $D_i$ participates,
which can lead to substantial over-clustering and a
failure to accurately model data in the presence of structural
independences between relations.
\Cref{fig:hirm-samples} illustrates and discusses this limitation
in further detail.

\subsection{Restrictions when Clustering Multiple Relations}
\label{sec:limitations-cluster-relations}

\begin{figure}[t]
\centering
\includegraphics[width=.8\linewidth]{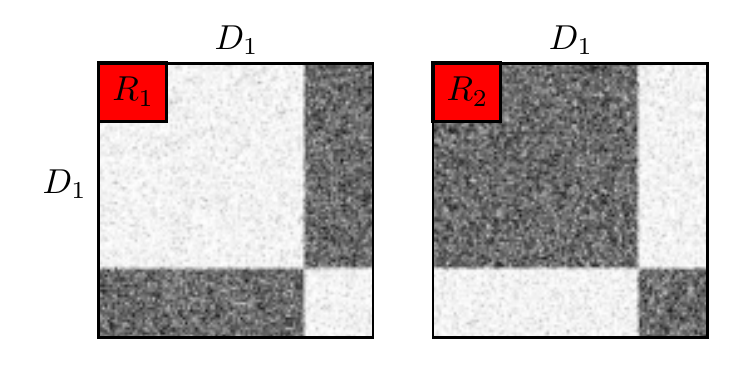}
\captionsetup{skip=0pt}
\caption{When used to cluster relations, the standard IRM~\citep{kemp2006}
uses a higher-ordering encoding that requires relations to be defined on the same domain
and assumes all relations within a cluster are identically distributed.
While $R_1$ and $R_2$ use identical partitions of $D_1$ and are anti-correlated, they are not
identically distributed and are thus assigned different
clusters by the IRM.
In contrast, the \hirm{} can learn clusters of relations defined on
different domains and can assign
non-identically distributed relations to the same cluster
(\cref{fig:nations-geopolitcs} shows a real-world example).}
\label{fig:irm-anti}
\end{figure}

\citet{kemp2006} applied the IRM to clustering multiple \textit{relations},
by treating the relations themselves as entities within a new domain.
More specifically, for a system with relations $R_1, \dots, R_m$,
all defined on same domain and codomain (say $D$ and $C$),
the key idea is to encode the system using one
higher-order relation $R': D' \times {D} \to C$, where the
entities of $D'$ are relations over $D$, i.e.,
$R'(j,i) \defas R_j(i)$ (for $1 \le j \le m$, $i \in D$).
While an IRM for $R'$ will learn a clustering of both $D'$ (the relations)
and $D$, there are at least two restrictions with this approach:
\begin{enumerate*}[label=(\roman*)]
\item\label{item:lim-domain} it only applies to relations defined on identical
  domains and codomains; and
\item\label{item:lim-iid} it clusters relations $R_i$ and $R_j$
  together only if they are both dependent and identically distributed
  (\cref{eq:irm-relation}).
\end{enumerate*}
\Cref{fig:irm-anti} illustrates and discusses this limitation in
further detail.


\section{Hierarchical Infinite Relational Model}
\label{sec:hirm}

We now present the \hirm{}, which addresses the aforesaid limitations
of the IRM by using a structure learning prior to infer probable
independences among relations that cannot be represented
structurally in a standard IRM.

Given a system $S$ with domains $D_1,\dots,D_n$ and relations
$R_1,\dots,R_m$, the \hirm{} first nonparametrically partitions the
$m$ \textit{relations} using a CRP (\cref{eq:crp-probs}),
where the cluster assignments of
the relations are denoted by $y \defas \set{y_1,\dots,y_m}$.
This partition induces a random number $K \defas \max\set{y_1,\dots,y_m}$
of subsystems $S_1,\dots,S_K$ of $S$.
For each $\ell=1,\dots,K$, the relations
  $\set{R_i \mid 1 \le i \le m, y_i = \ell}$ assigned to
  subsystem $S_\ell$ are modeled jointly by an IRM
  (\crefrange{eq:irm-crp}{eq:irm-relation}), independently of all
  relations assigned to another subsystem $S_{\ell'}$ ($\ell' \ne \ell$).
The \hirm{} thus defines a probability distribution over relation clusters,
domain entity clusters, and relation values with the following
factorization:
\begin{align}
\begin{aligned}[t]
&P(y_1,\dots,y_m, \set{z^{\ell 1}, \dots, z^{\ell n}}_{\ell=1}^{K}, R_1, \dots, R_m)
  \label{eq:hirm-factorization} \\
&\quad= P(y) \prod_{\ell=1}^{K}
    \prod_{i=1}^{n}P(z^{\ell i})
    \prod_{k \mid y_k = \ell}P(R_k \mid z^{\ell 1}, \dots, z^{\ell  n}).
    \hspace{-.5cm}
\end{aligned}
\end{align}
For each subsystem index $\ell = 1, \dots, K$; domain index
$i=1,\dots,n$; relation index $k=1,\dots,m$; entity indexes
$i_1,\dots,i_{t_k}$; and cluster indexes $j_1,\dots,j_{t_k}$, the
generative specification of the \hirm{} is given by the following process:
\begin{align}
\set{y_1, \dots, y_m} &\sim \crp(\gamma_0)
  \label{eq:hirm-crp-outer} \\
\set{z^{\ell i}_1,z^{\ell i}_2,\dots} &\sim \crp(\gamma_{\ell i})
  \label{eq:hirm-crp} \\
\theta_k(j_1,\dots,j_{t_k}) &\sim \pi_k(\lambda_k)
  \label{eq:hirm-parameter} \\
R_k(i_1,\dots,i_{t_k}) &\sim
  L_k(\theta_k({z^{y_k,d_{k1}}_{i_1}, \dots, z^{y_k,d_{kt_k}}_{i_{t_k}}})),
  \label{eq:hirm-relation}
\end{align}
where $(\gamma_0, \set{\set{\gamma_{\ell i}}_{i=1}^{n}}_{\ell =1}^{K}, \set{\lambda_k}_{k=1}^{m})$
are model hyperparameters, possibly endowed with their own hyperpriors.

The \hirm{} generalizes and extends the IRM.
First, it recovers the standard IRM when $\gamma_0 = 0$.
For $\gamma_0 > 0$, \cref{eq:hirm-crp-outer} specifies a CRP partition prior
over relations, where relations in the same block are modeled jointly using a
standard IRM (\crefrange{eq:hirm-crp}{eq:hirm-relation}).
In \cref{eq:hirm-crp}, each domain $D_i$ is associated with a
different partition $z^{\ell i}$ for each subsystem $S_\ell$ in which
it participates.
This inductive bias allows the \hirm{} to express
structural independences between relations and avoid modeling a Cartesian product of
domain partitions when the data for (a subset of) relations in the system are not well-aligned
(\cref{sec:limitations-overfit,fig:hirm-samples}).

Additionally, \cref{eq:hirm-crp-outer} allows the \hirm{} to
directly cluster dependent relations together, without using
higher-order encodings that are limited to relations defined on the
same domain as in the IRM (\cref{sec:limitations-cluster-relations}).
Further, \cref{eq:hirm-parameter,eq:hirm-relation} imply that
relations $R_k$ and $R_{k'}$ that are clustered together in a subsystem
$S_\ell$ need not be identically distributed (resp.\ \cref{fig:irm-anti}),
as they each have their
own parameters $\theta_{k}$ and $\theta_{k'}$, respectively.
The dependence is instead modeled by the shared domain partitions
$\set{z^{\ell 1}, \dots, z^{\ell n}}$ within subsystem $S_\ell$.
In sum, the nonparametric structure learning prior
\cref{eq:hirm-crp-outer} retains the benefits of the standard
IRM while addressing the limitations discussed in \cref{sec:limitations},
all within a Bayesian nonparametric model discovery framework.

%

\subsection{Posterior Inference}
\label{sec:hirm-inference}

An observed dataset $\set{r_1,\dots,r_m}$ for a relational system
consists of a finite number of realizations of relation values,
i.e., observations of random variables of the form
$\set{R_k(i_1,\dots,i_{t_k}) = r_k(i_1,\dots,i_{t_k})}$.
For notational ease and without loss of generality, we assume that the
relation values are fully observed for $N_i \ge 1$ entities
$\set{e^i_1,\dots,e^i_{N_i}}$ of each domain $D_i$ ($i=1,\dots,n$),
across all relations that it participates in (our reference
implementations of the \hirm{} handles arbitrary
index combinations with missing data).

Posterior inference in the \hirm{} is carried out by simulating an
ergodic Markov chain that converges to the distribution obtained by
conditioning \cref{eq:hirm-factorization} on the observed dataset.
%
%
The chain initializes a state $\mathscr{S}$ by sampling it from the prior
(\crefrange{eq:hirm-crp-outer}{eq:hirm-parameter}) and
iterates the state using Gibbs sampling.
\Cref{alg:mcmc-outer} shows one full Gibbs scan through all the variables
in the state $\mathscr{S}$.
We next describe transition operators for the updates in
\cref{algline:mcmc-y,algline:mcmc-z,algline:mcmc-theta}
of \cref{alg:mcmc-outer}.


\begin{algorithm}[t]
\algrenewcommand\algorithmicindent{1em}%
\caption{MCMC Gibbs Scan for \hirm{} (Sketch)}
\label{alg:mcmc-outer}
\begin{algorithmic}[1]
\Require Markov chain state $\mathscr{S}$ containing
  relation cluster assignments $\set{y_1,\dots,y_m}$,
  entity cluster assignments $\set{z^{\ell i}_{1}, \dots, z^{\ell i}_{N_i}}$,
  and parameters $\set{\theta_k(j_1,\dots,j_{t_k})}$,
  for $1 \le i \le m$, $1 \le \ell \le K$, and $1 \le k \le m$;
  dataset $r$.
\For{$k = 1, \dots, m$}
  \State \textbf{resample} $y_k$ given $(r, \mathscr{S} \setminus \set{y_k})$
  \label{algline:mcmc-y}
\EndFor
\For{$\ell = 1, \dots, \max(y_1,\dots,y_m)$}
  \LineComment{$I_\ell$ is set of domains in subsystem $S_\ell$}
  \State $I_\ell \gets \set{d_{kj} \mid 1 \le k \le m, 1 \le j \le t_k, y_k = \ell}$
  \For{$i \in I_{\ell}$}
    \For{$j = 1, \dots, N_i$}
      \State \textbf{resample} $z^{\ell i}_j$ given $(r, \mathscr{S} \setminus \set{z^{\ell,i}_j})$
      \label{algline:mcmc-z}
    \EndFor
  \EndFor
  \LineComment{$T_\ell$ is set of relations in subsystem $S_\ell$}
  \State $T_\ell  \gets \set{k\,{\mid}\, 1 {\le}\, {k}\,{\le} m, y_k\,{=}\,\ell, (\pi_k, L_k) \mbox{ nonconjugate}}$
  \For{$k \in T_{\ell}$}
    \For{$j_1 = 1, \dots, \max(z^{\ell d_{k1}}_1, \dots, z^{\ell d_{k1}}_{N_{d_{k1}}})$}
      \State \dots
      \For{$j_{t_k} = 1, \dots, \max(z^{\ell d_{kt_k}}_1, \dots, z^{\ell d_{kt_k}}_{N_{d_{kt_k}}})$}
        \State \textbf{resample} $\theta_k(j_1, \dots, j_{t_k})$ given $(r, \mathscr{S})$
          \label{algline:mcmc-theta}
      \EndFor
    \EndFor
  \EndFor
\EndFor
\end{algorithmic}
\end{algorithm}

\paragraph{Reampling relation cluster assignments $y_k$ \normalfont{:}}
This kernel uses the auxiliary Gibbs sampler~\citep[Algorithm 8]{neal2000}.
Let $C_{\ell} \defas \abs{\set{ k \mid 1 \le k \le K, y_k = \ell}}$
be the number of relations in $S_\ell$ and
$W_{\ell i} \defas \max\set{z^{\ell i}_{1}, \dots, z^{\ell i}_{N_i}}$
be the number of clusters for domain $D_i$ within $S_\ell$
($1 \le \ell \le K$).

\begin{enumerate}[label=Case~\arabic*:,wide=0pt]
\item If $y_k$ is a singleton ($C_k = 1$),
then it is resampled to take a new
value $\ell \in \set{1,\dots,K}$ with probability
\begin{align}
\begin{aligned}
  c_{k\ell}
  \prod_ {j_1=1}^{W_{\ell d_{k1\phantom{_k}}}}
    {\cdots}
    \prod_{j_{t_k}=1}^{W_{\ell d_{kt_k}}}
    w_{k\ell}(\mathbf{j}, \theta_k),
\end{aligned} \label{eq:weaselfish}
\end{align}
where $\mathbf{j} \defas (j_1,\dots,j_{t_k})$ and
\begin{align}
c_{k\ell} &\defas \begin{cases}
    {\gamma_0}/{(m-1+\gamma_0)} & \mbox{ if } \ell = y_k \\
    {C_\ell}/{(m-1+\gamma_0)} & \mbox{ otherwise}, \\
    \end{cases} \label{eq:rukh} \\
w_{k\ell}(\mathbf{j}, \theta_k)
  &\defas \prod_{\mathbf{i} \in A_{k\ell} (\mathbf{j})}
  L_k(r_k(\mathbf{i}); \theta_k(\mathbf{j})).
  \label{eq:ileocaecal}
\end{align}
\Cref{eq:rukh} is the conditional probability from the CRP prior
(\cref{eq:crp-probs}), and in \cref{eq:ileocaecal} the symbol
\begin{align}
A_{k\ell}(\mathbf{j})
  \defas \set{\mathbf{i} \mid
  z^{\ell d_{k1}}_{i_1} = j_{1}, \dots, z^{\ell d_{kt_k}}_{i_{t_k}} = j_{t_k}}
\end{align}
denotes the set of entity indexes $\mathbf{i} \defas (i_1,\dots,i_{t_k})$ for
domains $(d_{k1}, \dots, d_{kt_k})$ that are assigned to cluster
$\mathbf{j}$ of subsystem $S_\ell$
(where
  $1 \le k \le m$;
  $1 \le \ell \le M$;
  $1 \le j_1 \le W_{\ell d_{k1}}$;
  $\dots$;
  $1 \le j_{t_k} \le W_{\ell d_{kt_k}}$).
Note that if $(\pi_k, L_k)$ is a conjugate pair,
the parameters $\theta_k$ can be analytically
integrated out, and \cref{eq:ileocaecal} becomes
\begin{align}
w_{k\ell}(\mathbf{j}) \defas \int_{\theta}\Big[
    \prod_{\mathbf{i} \in A_{k\ell}(\mathbf{j})}
    L_k(r_k(\mathbf{i}); \theta)\Big]
  \pi_k(\theta; \lambda_k)d\theta.
\end{align}

\item If $y_k$ is not a singleton $(C_k > 1)$, then
\begin{enumerate}[label=\arabic*.]
\item For domain indexes $i=1,\dots,n$, draw cluster assignments
for a fresh entity partition, i.e.,
\begin{align}
\set{z^{K+1,i}_1, \dots, z^{K+1,i}_{N_i}} \sim \crp(\gamma), \\
W_{K+1,i} \defas \max \set{z^{K+1,i}_1, \dots z^{K+1,s}_{N_i}}.
\end{align}

\item Draw parameters
  $\theta_k(j_1,\dots,j_{t_k})$
  for relation indexes $k=1,\dots,m$
  and cluster indexes
  $j_1 = 1,\dots, W_{K+1,d_{k1}}$;
  $\dots$;
  $j_{t_k} = 1,\dots, W_{K+1,d_{kt_k}}$.
\end{enumerate}

Next, resample $y_k$ to take a new value $\ell \in \set{1,\dots,K+1}$
using the same terms in \crefrange{eq:weaselfish}{eq:ileocaecal}
from the previous case, except that the CRP weight
$c_{k\ell}$ in \cref{eq:rukh} is instead
\begin{align}
\hspace{-.225cm}
c_{k\ell} \defas \begin{cases}
  {(C_\ell - 1)}/{(m-1+\gamma_0)} & \mbox{ if } \ell = y_k \\
  {C_\ell}/{(m-1+\gamma_0)} & \mbox{ if } \ell \ne y_k, \ell \le K \\
  {\gamma_0}/{(m-1+\gamma_0)} & \mbox{ if }  \ell = K + 1.
\end{cases}
\end{align}
\end{enumerate}

\paragraph{Resampling entity cluster assignments $z^{\ell i}_{j}$\normalfont{:}}
Within each subsystem $S_\ell$, the entity cluster
assignments are transitioned using the collapsed Gibbs sampler~\citep[Alg.~3]{neal2000}.
Alternatively, the split-merge algorithm can be used~\citep{jain2004}.
\citet{xu2007} discuss additional sampling-based and variational approaches for
these variables.

\paragraph{Resampling cluster parameters $\theta_k(j_1,\dots,j_{t_k})$\normalfont{:}}
Sample $\theta_k'(\mathbf{j}) \sim q_k(\theta_k(\mathbf{j}))$
from a proposal distribution (e.g., the prior $\pi_k(\lambda_k)$ or Gaussian drift
$\mathcal{N}(\theta_k(\mathbf{j}), \sigma_k)$)
and accept the move according to the Metropolis-Hastings probability
\begin{align}
\min\left(1, \frac{
  \pi_k(\theta'_k(\mathbf{j}); \lambda_k)
  w_{k\ell}(\mathbf{j}, \theta'_k)
    q_k(\theta_k(\mathbf{j}); \theta'_k(\mathbf{j}))
}{
  \pi_k(\theta_k(\mathbf{j}); \lambda_k)
  w_{k\ell}(\mathbf{j}, \theta_k)
    q_k(\theta'_k(\mathbf{j}); \theta_k(\mathbf{j}))
}\right).
\end{align}
where is $w_{k\ell}$ (\cref{eq:ileocaecal}) is the data likelihood for cluster $\mathbf{j}$.

\paragraph{Resampling hyperparameters\normalfont{:}}
Broad exponential hyperpriors are used for all the model hyperparameters
$\gamma_0, \set{\gamma_{\ell i}}, \set{\lambda_k}$
that appear in \crefrange{eq:hirm-crp-outer}{eq:hirm-parameter},
which are resampled using gridded-Gibbs~\citep{ritter1992}.
It is also possible to instead use slice sampling~\citep{neal2003}.

\section{Evaluation}
\label{sec:evaluations}
We implemented a prototype of the \hirm{}%
\footnote{Reference implementations of the \hirm{} in C++ and Python
  are available at \url{https://github.com/probcomp/hierarchical-irm}.}
and evaluated it in three settings:
  solving density estimation tasks in object-attribute data;
  discovering relational structure in political data; and
  learning relationships between gene properties.

\subsection{Object-Attribute Benchmarks}
\label{sec:evaluations-object-attribute}


\begin{table}[t]
\newcommand{\win}{\bullet}
\newcommand{\los}{\circ}
\newcommand{\tie}{\phantom{\bullet}}
\captionsetup{skip=0pt, font=footnotesize}
\caption{Prediction accuracy of \hirm{} and Bayesian nonparametric
baselines on a benchmark of 20 object-attribute datasets.}
\label{table:binary}
\begin{adjustbox}{max width=\linewidth}
\begin{tabular}{|l||c|c|c||c|c|c|}
\hline
~            & \multicolumn{3}{c||}{\bfseries Dataset Statistics} & \multicolumn{3}{c|}{\bfseries Average Test Log-Likelihood} \\ \hline
Dataset      & $N_{\rm cols}$ & $N^{\rm train}_{\rm rows}$ &$N^{\rm test}_{\rm rows}$ & \hirm{} & IRM & DPMM \\ \hline\hline
NLTCS        & 16   & 18338  & 3236  & -06.00  & -06.01  $\win$ & -06.01 $\win$  \\
MSNBC        & 17   & 330212 & 58265 & -06.19  & -06.27  $\win$ & -06.22 $\win$  \\
KDDCup 2000  & 64   & 199999 & 34955 & -02.13  & -02.13  $\win$ & -02.13 $\win$  \\
Plants       & 69   & 19733  & 3482  & -13.75  & -14.23  $\win$ & -13.81 $\tie$  \\
Audio        & 100  & 17000  & 3000  & -39.99  & -40.34  $\tie$ & -40.02 $\tie$  \\ \hline
Jester       & 100  & 10000  & 4116  & -52.91  & -52.96  $\tie$ & -52.92 $\tie$  \\
Netflix      & 100  & 3500   & 3000  & -56.96  & -57.48  $\win$ & -56.96 $\tie$  \\
Accidents    & 111  & 14458  & 2551  & -33.85  & -39.43  $\win$ & -38.93 $\win$  \\
Retail       & 135  & 24979  & 4408  & -10.90  & -10.99  $\tie$ & -10.92 $\tie$  \\
Pumsb-star   & 163  & 13897  & 2452  & -32.77  & -38.95  $\win$ & -38.02 $\win$  \\ \hline
DNA          & 180  & 2000   & 1186  & -87.65  & -97.44  $\win$ & -97.62 $\win$  \\
Kosarek      & 190  & 37825  & 6675  & -10.91  & -10.99  $\tie$ & -10.95 $\tie$  \\
MSWeb        & 294  & 62191  & 5000  & -10.23  & -11.20  $\win$ & -10.26 $\tie$  \\
Book         & 500  & 9859   & 1739  & -34.43  & -34.52  $\tie$ & -34.76 $\tie$  \\
EachMovie    & 500  & 5526   & 591   & -52.23  & -52.09  $\tie$ & -54.86 $\tie$  \\ \hline
WebKB        & 839  & 3361   & 838   & -156.67 & -157.27 $\tie$ & -158.26$\tie$  \\
Reuters-52   & 889  & 7560   & 1540  & -90.22  & -90.06  $\tie$ & -89.34 $\tie$  \\
20 Newsgroup & 910  & 15057  & 3764  & -153.52 & -156.46 $\win$ & -153.95$\tie$  \\
BBC          & 1058 & 1895   & 330   & -253.36 & -253.86 $\tie$ & -254.59$\tie$  \\
Ad           & 1556 & 2788   & 491   & -45.19  & -46.17  $\tie$ & -52.40 $\win$  \\
\hline
\multicolumn{7}{l}{$\win$ indicates significantly worse than \hirm{} ($p=0.05$ Mann-Whitney U test).}
\end{tabular}
\end{adjustbox}
\end{table}


\begin{table}[b]
\captionsetup{skip=0pt, font=footnotesize}
\caption{Summary of no.~of wins, ties, and losses of \hirm{} on benchmarks
from \cref{table:binary}, compared to two Bayesian
nonparametric baselines and two probabilistic deep learning baselines.}
\label{table:challenge}
\begin{adjustbox}{max width=\linewidth}
\begin{tabular}{|l||llll|}
\hline
\multicolumn{1}{c}{~} & IRM & DPMM & LearnSPN & RAT-SPN \\ \hline\hline
\hirm{} \# win & 11 & 7 & 6 & 4 \\
\hirm{} \# tie & 9 & 13 & 8 & 13\\
\hirm{} \# lose & 0 & 0 & 6 & 3 \\ \hline\hline
\end{tabular}
\end{adjustbox}
\end{table}

\begin{figure}[t]
\begin{subfigure}[t]{.49\linewidth}
\includegraphics[width=\linewidth]{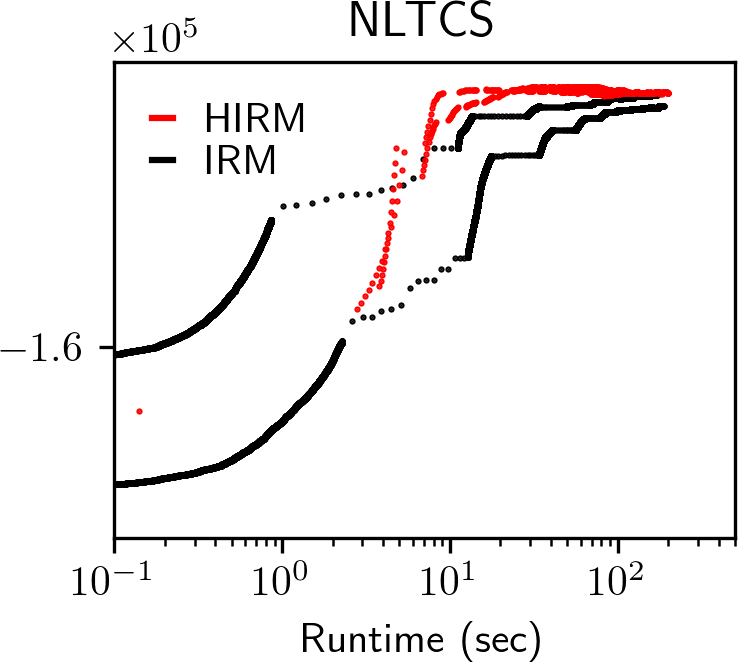}
\end{subfigure}\hfill%
\begin{subfigure}[t]{.49\linewidth}
\includegraphics[width=\linewidth]{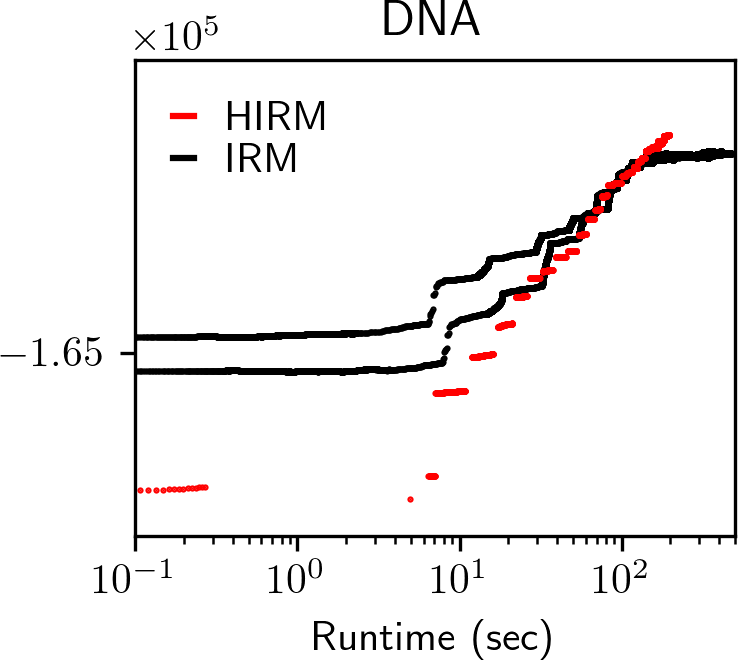}
\end{subfigure}
\begin{subfigure}[t]{.49\linewidth}
\includegraphics[width=\linewidth]{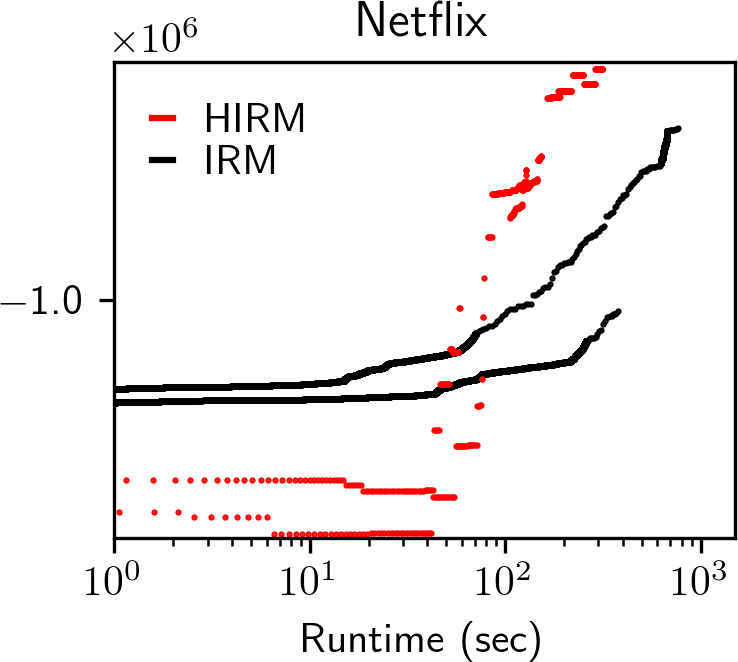}
\end{subfigure}\hfill%
\begin{subfigure}[t]{.49\linewidth}
\includegraphics[width=\linewidth]{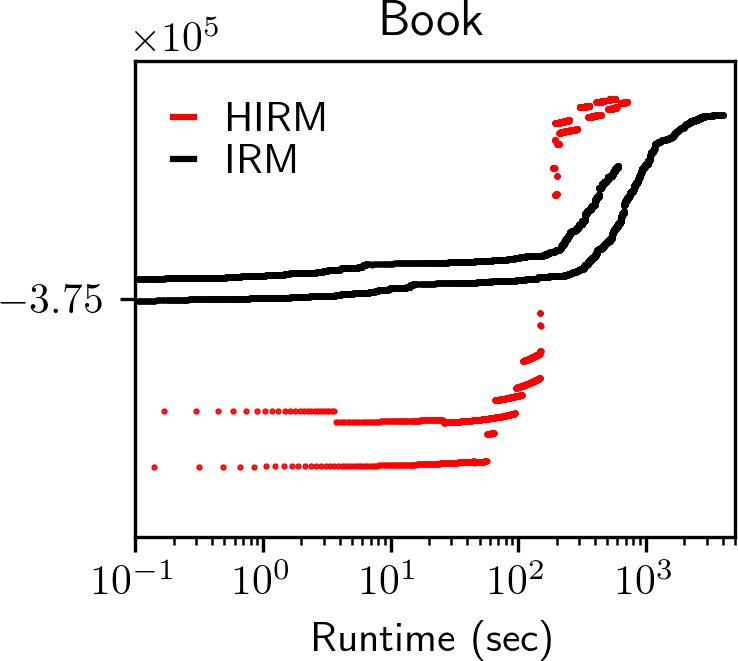}
\end{subfigure}%
\captionsetup{belowskip=-10pt,aboveskip=5pt}
\caption{Runtime vs.~held-in data log score for \hirm{} and IRM in
four representative benchmarks from \cref{table:binary}.
For each method, two independent runs of inference are plotted.}
\label{fig:runtime}
\end{figure}

We assessed the predictive performance of the \hirm{} on a benchmark
of 20 object-attribute datasets~\citep{gens2013} and compared the
results to two Bayesian nonparametric baselines.
In \cref{table:binary}, the first four columns summarize the dataset statistics
(16--1556 columns, 2000--330212 rows).
The last three columns show the test log-likelihood from the
\hirm{}, IRM~\citep{kemp2006,xu2006}, and Dirichlet process mixture
model~\citep[DPMM;][]{lo1984}.
As in~\citet{kemp2006}, the IRM encodes object-attribute data using one
binary relation $R: \mathrm{Attr} \times \mathrm{Obj} \to \set{0,1}$.
The \hirm{} encodes each dataset using $N_{\rm cols}$
unary relations $\set{R_i: \mathrm{Obj} \to \set{0,1} \mid i \in
\mathrm{Attr}}$ with structure learning
(\cref{eq:hirm-crp-outer}) over the dependence between the
attributes.
The DPMM uses the same encoding as the \hirm{} but without structure
learning (i.e., all attributes are modeled jointly).
Dots indicate significantly worse values than the \hirm{}
($p\,{=}\,0.05$, Mann--Whitney U test on the $N^{\rm test}_{\rm rows}$
predictions from each model).
\Cref{table:binary} shows that the \hirm{} consistently
outperforms these baselines---it is significantly better in 17 cases
and worse in zero cases.
\Cref{fig:runtime} shows a plot of runtime vs.\ held-in data log score
for two runs of the \hirm{} and IRM on four of the benchmarks.
Despite using a structure learning prior, the runtime of the \hirm{}
matches or outperforms the IRM; in fact, the \hirm{} often infers
simpler partitions within the independent subsystems, which can
improve both the runtime scaling and model fit.

To further assess the density estimation results, we compared the
\hirm{} test log-likelihood to those obtained from probabilistic deep
learning baselines for object-attribute data:
LearnSPN~\citep{gens2013} and RAT-SPN~\citep{peharz2019}.
\Cref{table:challenge} summarizes the comparison
(tie means statistically insignificant differences).
The results show that the \hirm{}, which is a relatively shallow Bayesian
model (\crefrange{eq:hirm-crp-outer}{eq:hirm-relation}), is
competitive on object-attribute data with higher capacity
probabilistic deep learning baselines that fit the data using greedy
search.
The \hirm{} is distinguished by being additionally applicable to far
more general relational systems, as we next demonstrate.


\begin{figure*}[t]

\centering

\begin{subtable}{\linewidth}
\centering
\captionsetup{skip=0pt, textfont={normalsize}}
\caption{Subset of countries (15 total), attributes (111 total), and interactions (56 total) in the relational system.}
\label{table:nations}
\begin{tabularx}{\linewidth}{|X|l|l@{}|}
\hline\hline
Countries (Domain) &  Indonesia, Jordan, Burma, India, Israel, Egypt, Poland, USSR, UK, USA, Brazil \\ \hline
Attributes (Unary Relations) & Area, Telephone Users, Communist, Literacy, Protests, Purges, Democracy, \dots\\ \hline
Interactions (Binary Relations) & Exports, Enemies, Allies, Economic Aid, Book Translations, Treaties, Tourism, \dots \\
\hline\hline
\end{tabularx}
\hrule
\end{subtable}

\captionsetup[subfigure]{skip=-4pt}

\begin{subfigure}{.95\linewidth}
\begin{subfigure}[b]{.25\linewidth}
\includegraphics[width=\linewidth]{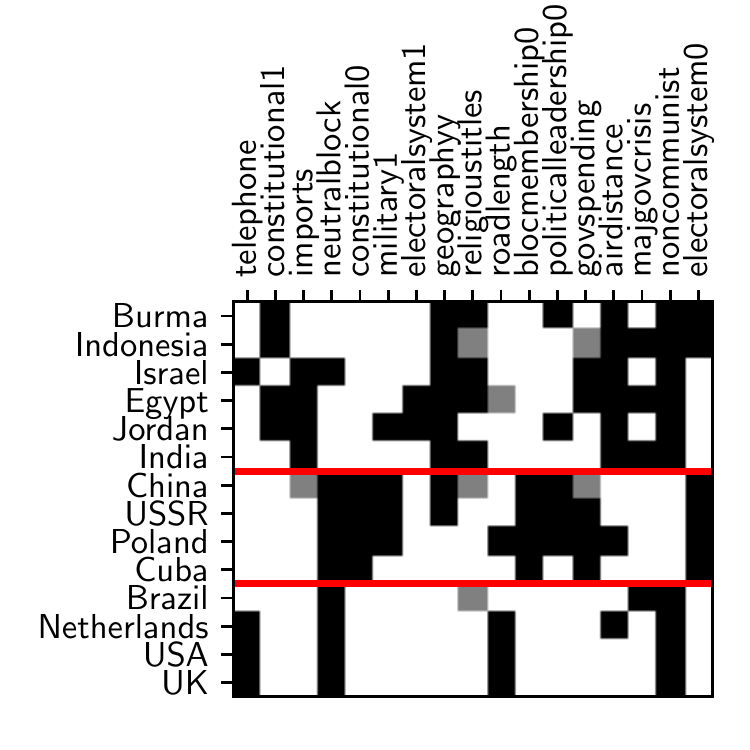}%
\caption*{Attributes}
\end{subfigure}%
\begin{subfigure}[b]{.25\linewidth}
\includegraphics[width=\linewidth]{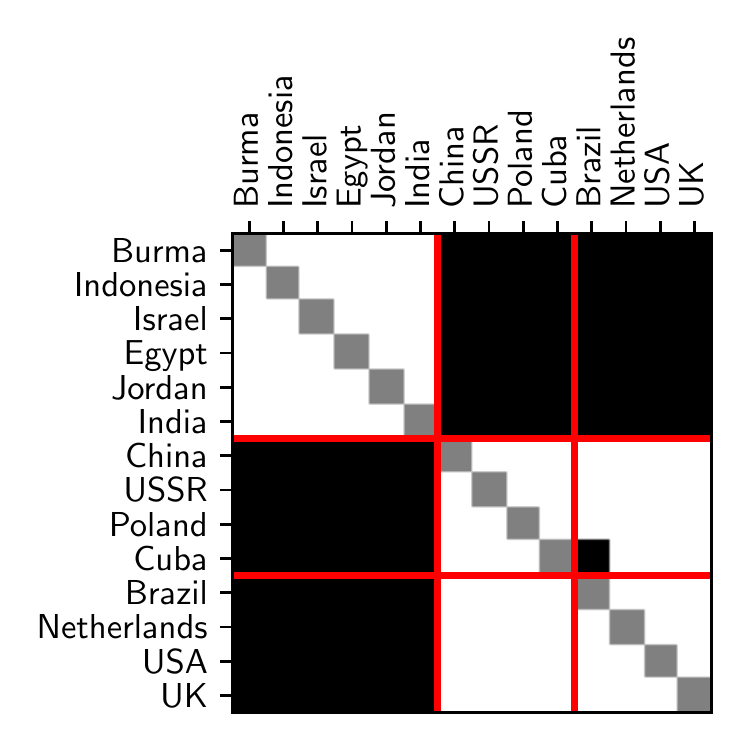}%
\caption*{Neutral}
\end{subfigure}%
\begin{subfigure}[b]{.25\linewidth}
\includegraphics[width=\linewidth]{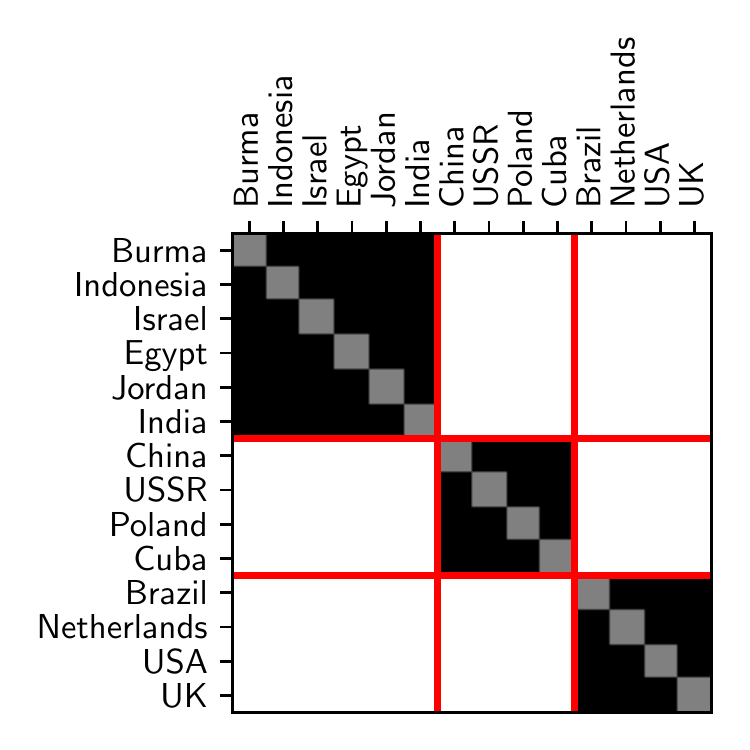}%
\caption*{Allies}
\end{subfigure}%
\begin{subfigure}[b]{.25\linewidth}
\includegraphics[width=\linewidth]{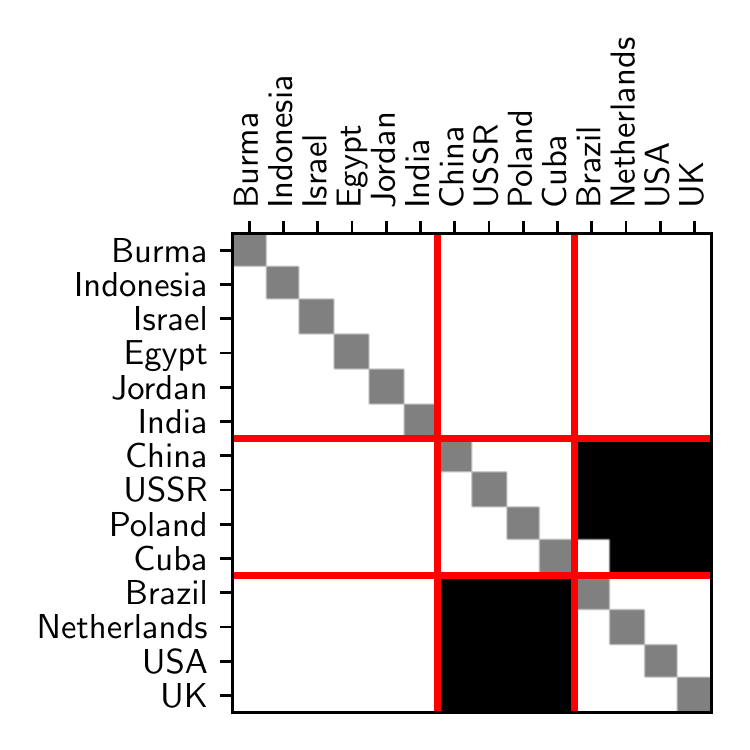}%
\caption*{Rivals}
\end{subfigure}
  \captionsetup{skip=0pt,textfont={normalsize},}
  \caption{Inferred Subsystem 1 (Geopolitical Blocs)}
  \label{fig:nations-geopolitcs}
\end{subfigure}
\hrule\hrule

\begin{subfigure}{.95\linewidth}
\begin{subfigure}[b]{.25\linewidth}
\includegraphics[width=\linewidth]{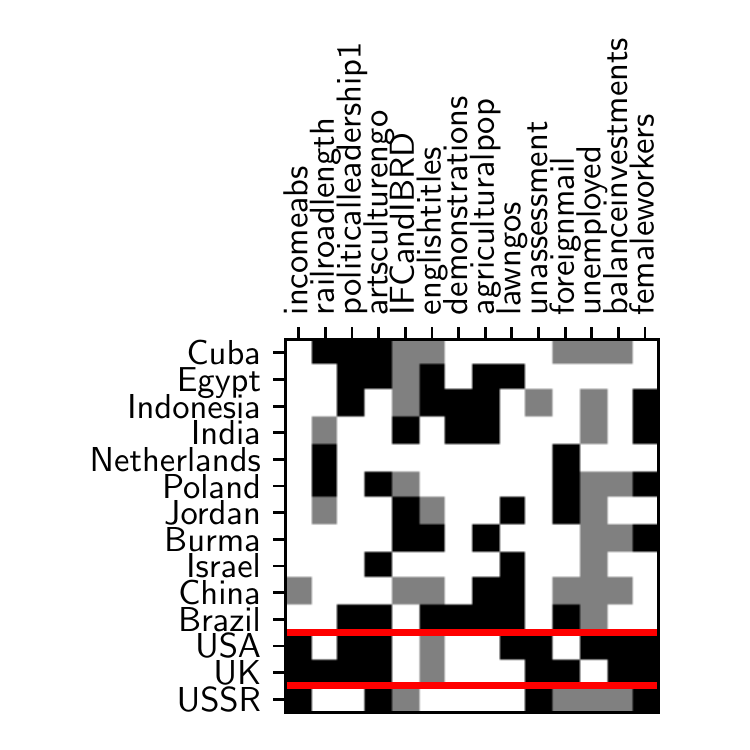}%
\caption*{Attributes}
\end{subfigure}%
\begin{subfigure}[b]{.25\linewidth}
\includegraphics[width=\linewidth]{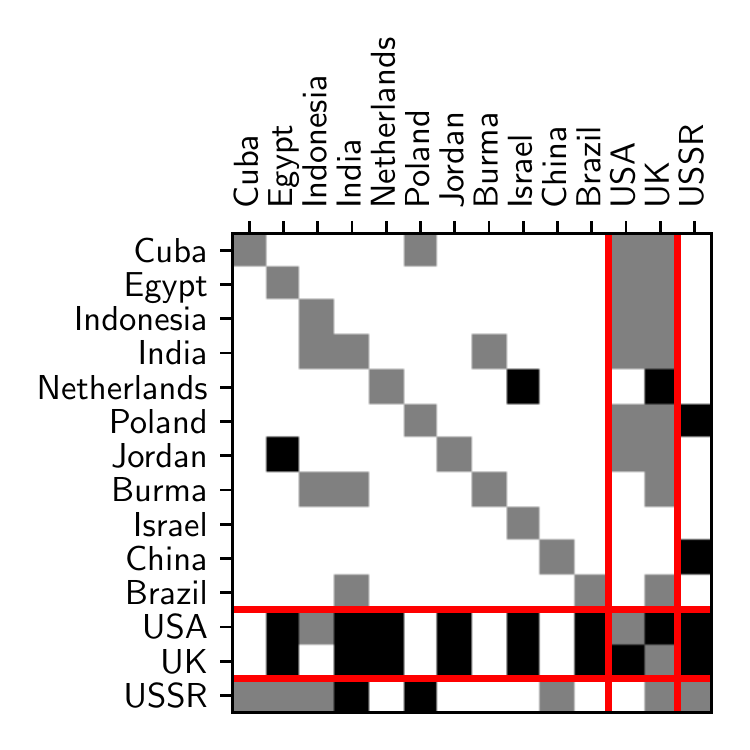}%
\caption*{Tourism}
\end{subfigure}%
\begin{subfigure}[b]{.25\linewidth}
\includegraphics[width=\linewidth]{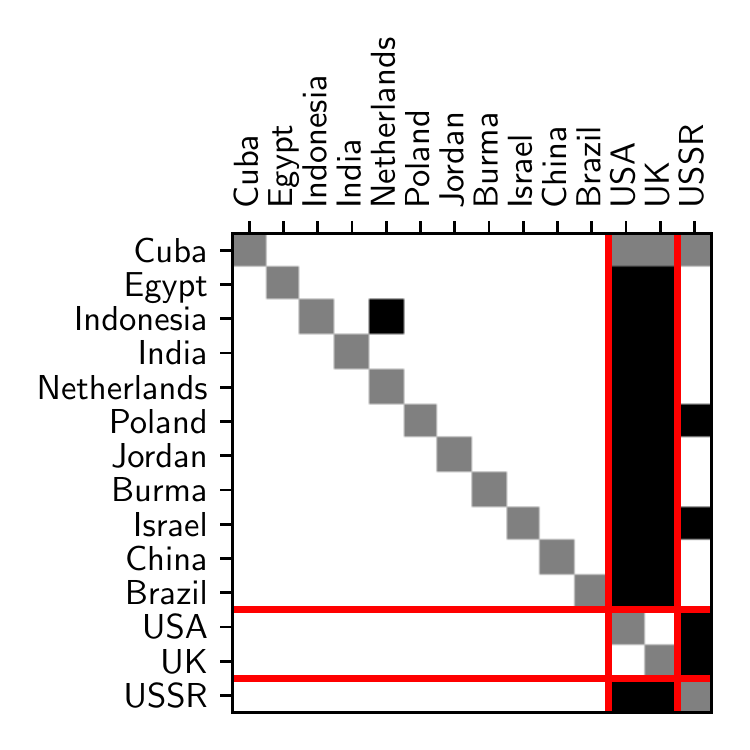}%
\caption*{Book Translations}
\end{subfigure}%
\begin{subfigure}[b]{.25\linewidth}
\includegraphics[width=\linewidth]{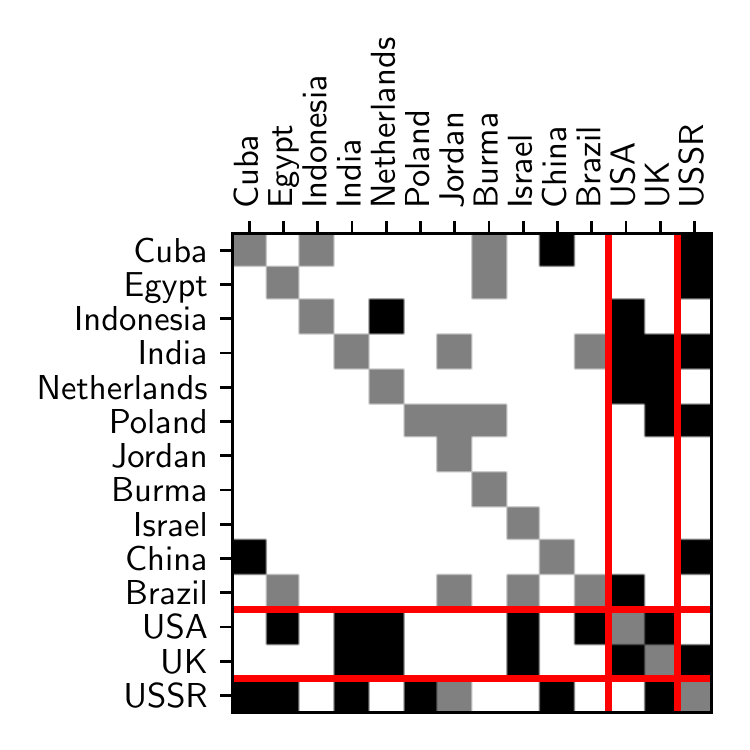}%
\caption*{Exports}
\end{subfigure}
\captionsetup{skip=0pt,belowskip=0pt,textfont={normalsize}}
\caption{Inferred Subsystem 2 (Economy and Culture)}
\label{fig:nations-books}
\end{subfigure}

\hrule\hrule

\begin{subfigure}{\linewidth}
\centering
\begin{subfigure}[b]{.475\linewidth}
\begin{subfigure}{.5\linewidth}
\includegraphics[width=\linewidth]{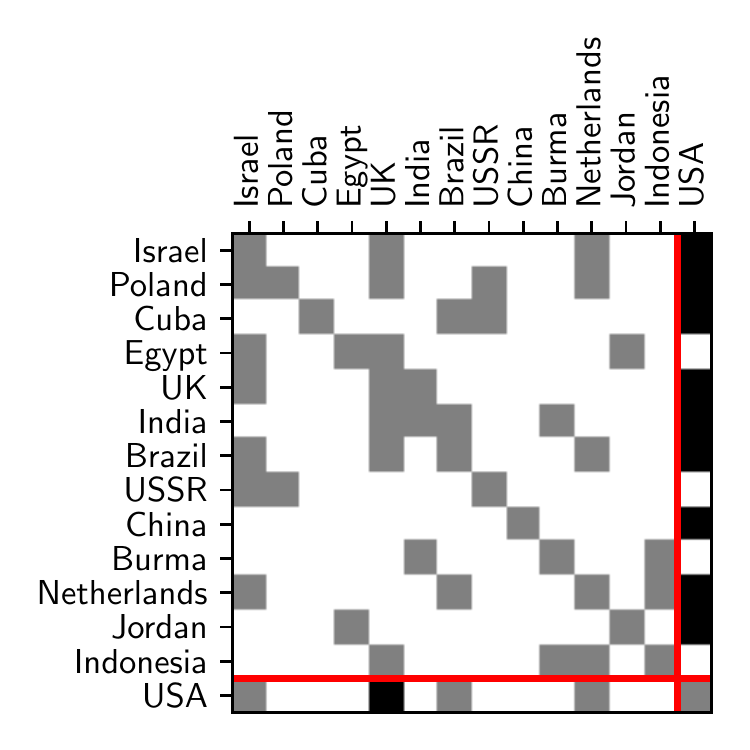}%
\caption*{Emigration}
\end{subfigure}%
\begin{subfigure}{.5\linewidth}
\includegraphics[width=\linewidth]{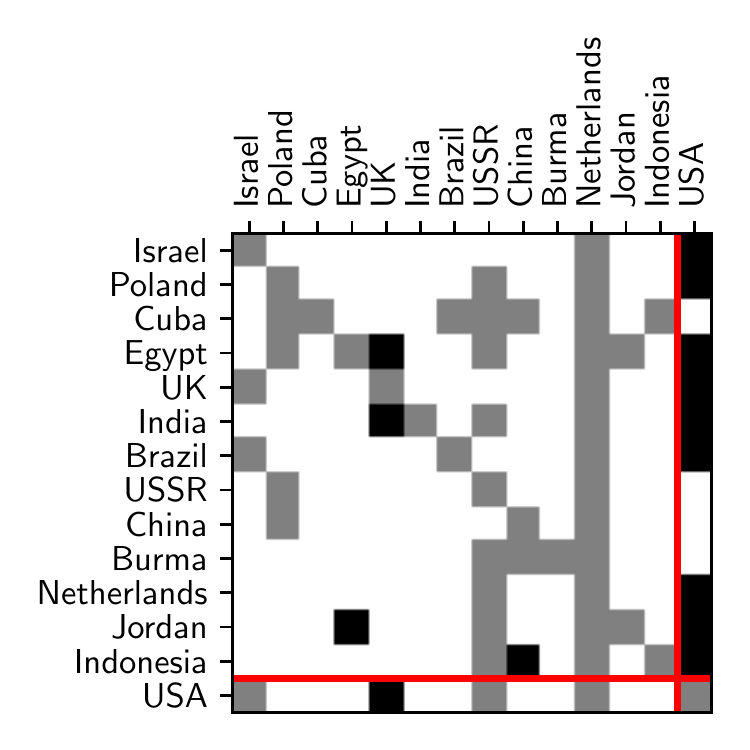}%
\caption*{Students}
\end{subfigure}%
\captionsetup{skip=0pt,belowskip=0pt,textfont={normalsize}}
\caption{Inferred Subsystem 3 (USA Outlier)}
\label{fig:nations-outlier}
\end{subfigure}\vrule\vrule
\begin{subfigure}[b]{.475\linewidth}
\begin{subfigure}{.5\linewidth}
\includegraphics[width=\linewidth]{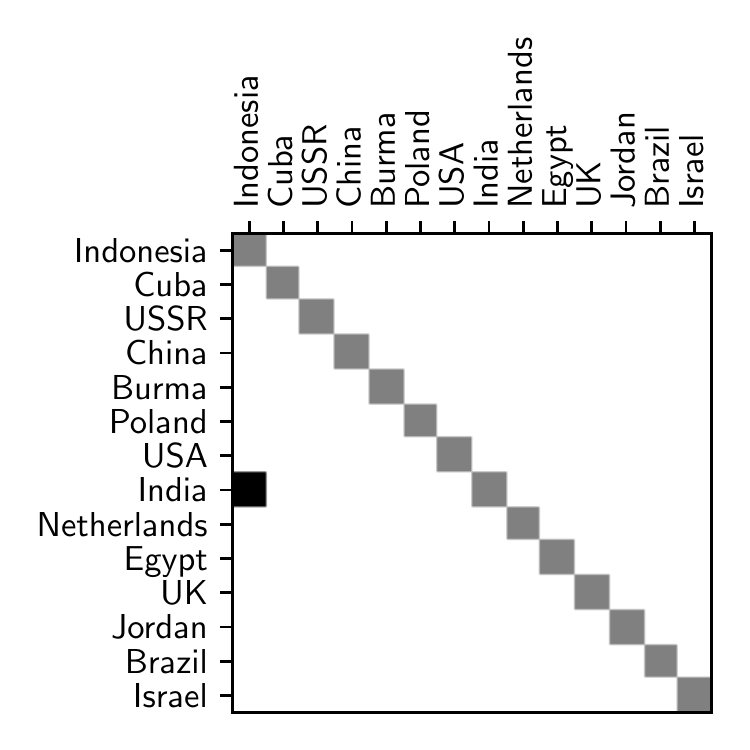}%
\caption*{Attack Embassy}
\end{subfigure}%
\begin{subfigure}{.5\linewidth}
\includegraphics[width=\linewidth]{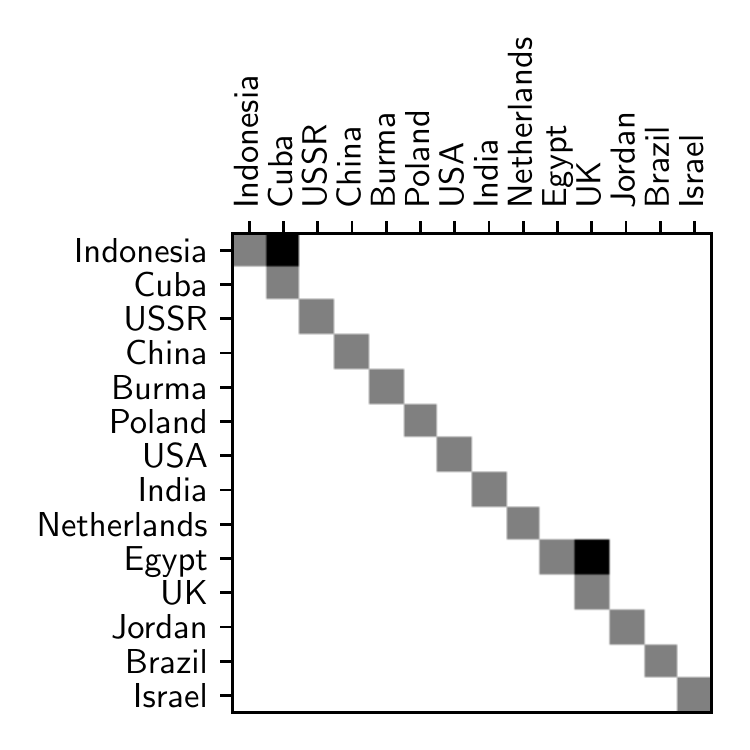}%
\caption*{Sever Relations}
\end{subfigure}%
\captionsetup{skip=0pt,belowskip=0pt,textfont={normalsize}}
\caption{Inferred Subsystem 4 (Sparse Interactions)}
\label{fig:nations-sparse}
\end{subfigure}
\hrule\hrule
\end{subfigure}

\captionsetup{belowskip=-10pt, aboveskip=2pt}
\caption{Systems of concepts inferred by the \hirm{} on the ``Dimensionality of Nations''
data (schema in \cref{fig:system-nations}).}
\label{fig:nations}
\end{figure*}


\begin{figure*}
\begin{subfigure}[t]{.17\linewidth}
\captionsetup{skip=1pt}
\caption{Inferred \domain{Gene} Clusters~{(for two contexts)}}
\label{fig:genes-cluster-gene}
\begin{tikzpicture}
\node[
  draw=black,
  inner sep=2pt,
  label={[yshift=-1.25cm,xshift=.12cm,anchor=west,fill=white,inner sep=1pt]left:{\footnotesize At \domain{Localization}}},
  ]{\includegraphics[width=\textwidth]{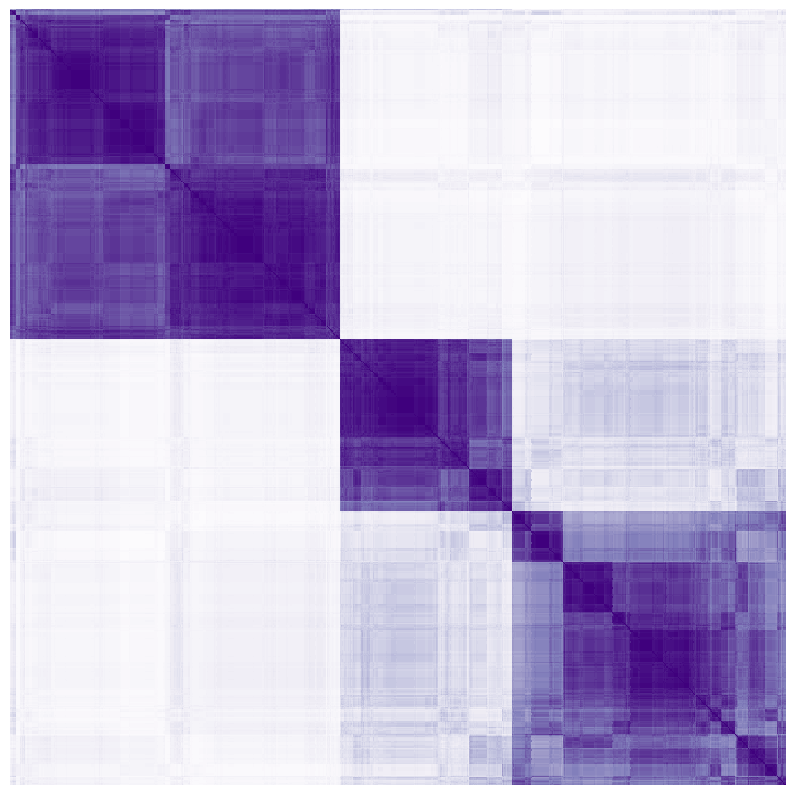}};
\end{tikzpicture}
\begin{tikzpicture}
\node[
  draw=black,
  inner sep=2pt,
  label={[yshift=-1.25cm,xshift=.12cm,anchor=west,fill=white,inner sep=1pt]left:{\footnotesize Belong \domain{Class}}},
  ]{\includegraphics[width=\textwidth]{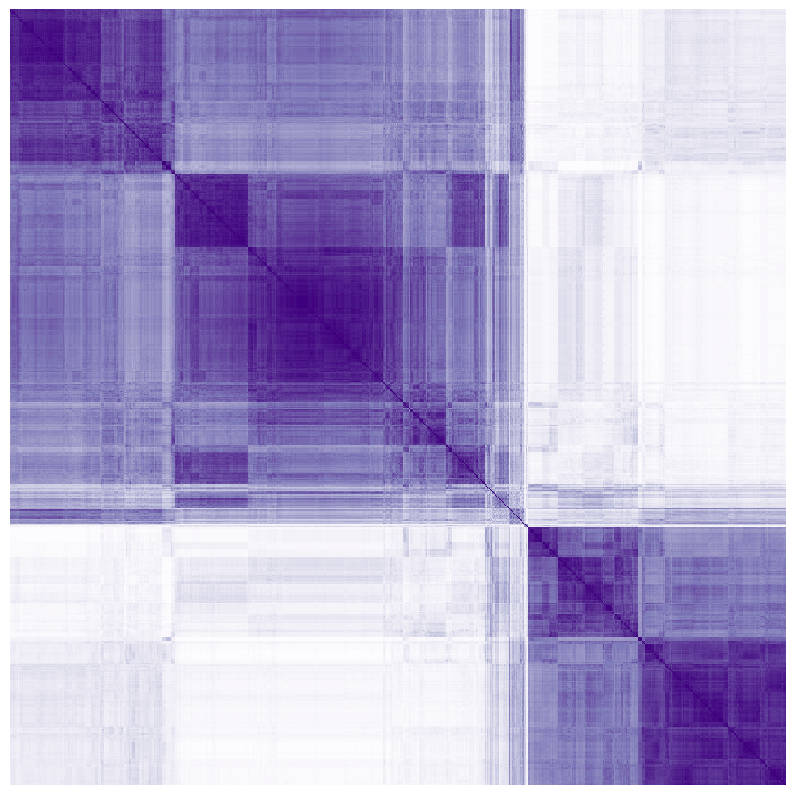}};
\end{tikzpicture}
\end{subfigure}\hfill%
\begin{subfigure}[t]{.375\linewidth}
\captionsetup{skip=0pt}
\caption{Inferred \domain{Localization} Clusters}
\label{fig:genes-cluster-localization}
\setlength{\FrameSep}{2pt}
\begin{framed}
\includegraphics[width=\textwidth]{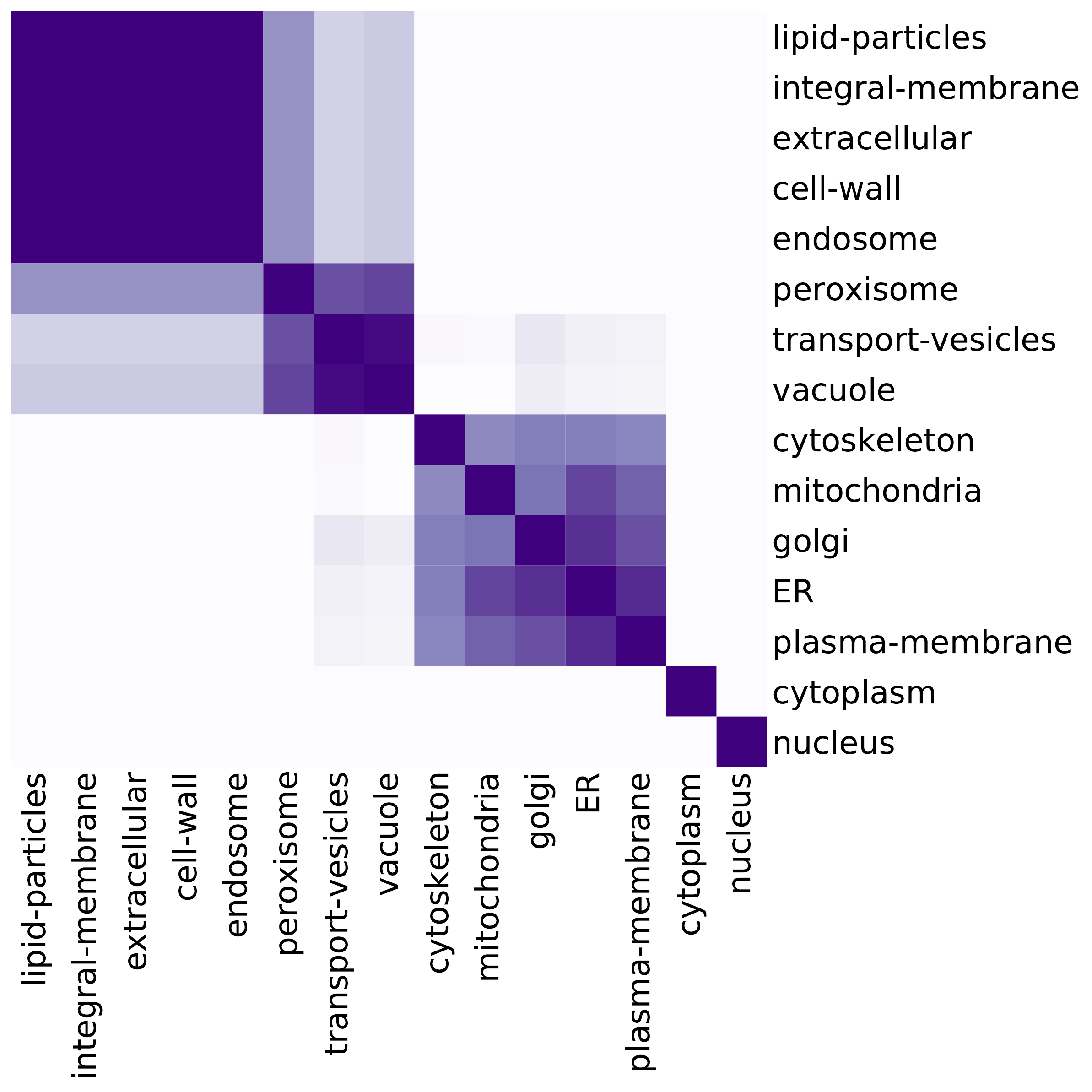}
\end{framed}
\end{subfigure}\qquad%
\begin{subfigure}[t]{.375\linewidth}
\captionsetup{skip=0pt}
\caption{Inferred \domain{Class} Clusters}
\label{fig:genes-cluster-class}
\begin{framed}
\includegraphics[width=\textwidth]{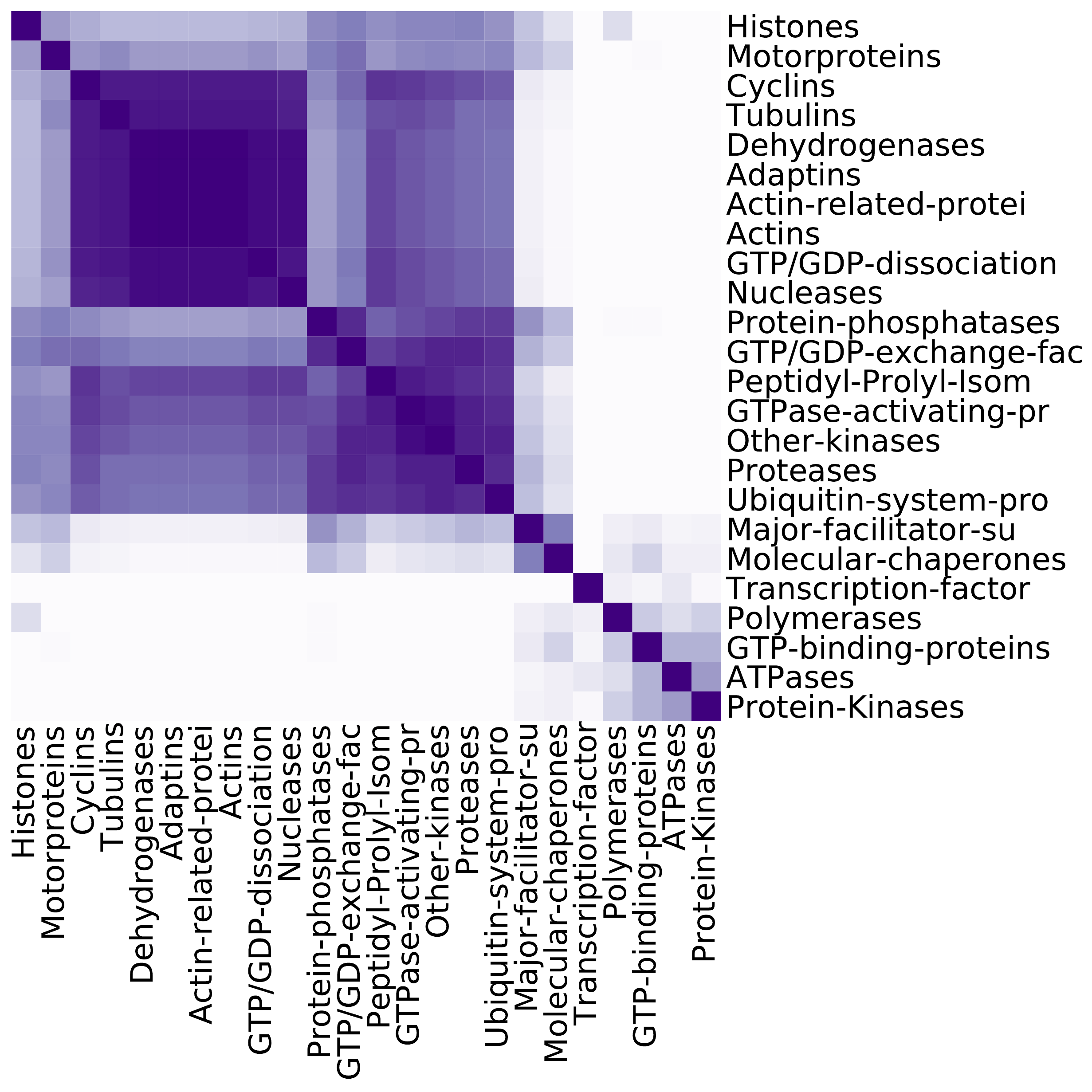}
\end{framed}
\end{subfigure}
\captionsetup{belowskip=-10pt,skip=5pt}
\caption{Posterior co-clustering probabilities for various
relational domains in yeast genome data (schema in \cref{fig:system-genes}).}
\label{fig:genes}
\end{figure*}

\subsection{Political Interactions}
\label{sec:evaluations-political}

We next applied the \hirm{} to the ``Dimensionality of Nations''
project~\citep{rummel1999}, using the version dataset from~\citet{kemp2006}
for years 1950--1965.
\Cref{fig:system-nations,table:nations} show a subset of the 15
countries, 111 attributes and 56 interactions.
\crefrange{fig:nations-geopolitcs}{fig:nations-sparse} show a
collection of independent subsystems of relations discovered by the
\hirm{} (gray cells indicate missing values).
Each inferred subsystem reflects a different partition of the
countries that explains the attribute and interactions within
the subsytem.
For example, in \cref{fig:nations-geopolitcs}, the \hirm{} finds that the
geopolitcal bloc interactions are associated with attributes such as ``electoral
system'', ``political leadership'', and ``constitutional''.\footnote{%
In \cref{fig:nations-geopolitcs}, the Cuba--Brazil relationship is neutral
despite the countries belonging to rival geopolitcal
blocs, which is detected by the \hirm{} as probabilistically unlikely.
This outlier is explained by the so-called the
American--Brazilian--Cuban ``triangular diplomacy''
during the 1962 missile crisis~\citep{hershberg2004}.}
In \cref{fig:nations-books}, which represents economic and cultural
ties and includes attributes such as ``absolute income'', ``agricultural
population'', and ``arts and culture NGOs'', the data shows that
tourists from the UK and USA travel to countries from all clusters
and all countries translate books from the USA and UK, who in turn
translate books from the USSR.
\Cref{fig:nations-outlier} represents a subsystem of relations in
which the USA is a clear outlier due to its unusually high number of
immigrants and foreign students: the \hirm{} has inferred that these
interactions are independent of the fact that China and Russia, for example,
are geopolitical rivals of the USA (\cref{fig:nations-geopolitcs}).
\Cref{fig:nations-sparse} contains sparse relations such
as ``Attack Embassy'' and ``Sever Relations'', which
form a subsystem with one country cluster and a
small probability for the hostile event.

In contrast to the \hirm{}, the IRM cannot detect subsystem structure
of this form since it uses a single country partition for all
interactions, which is an inaccurate explanation of the data in light of the widely
varying interaction patterns in the subsystems
(\crefrange{fig:nations-geopolitcs}{fig:nations-sparse})
discovered by the \hirm{}.

\subsection{Genomic Properties}
\label{sec:evaluations-genomics}

Our third application of the \hirm{} is to structure discovery in a
widely used dataset of yeast genomes~\citep{cheng2001}.
\Cref{fig:system-genes} shows a diagram of the relational
system.
There are nine domains: the \domain{Gene} domain has 1,243 unique
identifiers and the remaining domains represent gene properties.
There is one binary relation between \domain{Gene} and each of the
eight other domains, as well as one binary relation (Interact) on
\domain{Gene}.
A single gene is typically involved in multiple relations
with the \domain{Complex}, \domain{Phenotype},
\domain{Class}, \domain{Motif}, and \domain{Function} domains, but has
only one value for \domain{Essential} and \domain{Chromosome}.
\Cref{table:gene} shows an example record for gene G235131:
some characteristics of this gene are that the \domain{Class} is
missing, it forms two \domain{Complex}, has two \domain{Function};
there are five observed \domain{Phenotype}; and it interacts with 11
other genes (three of which are listed).

\begin{table}[t]
\captionsetup{skip=0pt}
\caption{Example Gene}
\label{table:gene}
\footnotesize
\begin{tabularx}{\linewidth}{|l|X|}
\hline
\multicolumn{1}{|c|}{\bfseries Field} & \multicolumn{1}{c|}{\bfseries Value} \\ \hline
\domain{Gene}         & G235131 \\
\domain{Essential}    & Non-Essential \\
\domain{Class}        & ? \\
\domain{Complex}      & Histone Acetyltransferase \\
\quad ---             & Transcriptosome \\
\domain{Phenotype}    & Auxotrophies \\
\quad ---             & Carbohydrate \& Lipid Biosynth. \\
\quad ---             & Conditional Phenotypes \\
\quad ---             & Mating \& Sporulation Defects \\
\quad ---             & Nucleic Acid Metab.~Defects \\
\domain{Motif}        & PS00633 \\
\domain{Chromosome}   & 2 \\
\domain{Function}     & Transcription \\
\quad ---             & Cellular Organization \\
\domain{Localization} & Nucleus \\
Interactions          & G234980, G235780, G235278, \dots \\ \hline
\end{tabularx}
\vspace{-.5cm}
\end{table}

\Cref{fig:genes-cluster-gene} shows two heatmaps that summarize
the clusterings of genes learned by the \hirm{} under two different
contexts.
More specifically, each row and column in a heatmap represents a
unique \domain{Gene} and the color of a cell is the posterior
probability (between 0 and 1) that the two genes are assigned to the
same latent cluster (estimated by an ensemble of 100 posterior \hirm{} samples).
The top (resp.~bottom) heatmap in \cref{fig:genes-cluster-gene}
shows posterior co-clustering probabilities conditioned on being in
the subsystem that contains the ``\domain{Gene} At \domain{Localization}''
(resp. ``\domain{Gene} Belong \domain{Class}'') relation, which we call a ``context''.
These heatmaps reflect a key feature of the \hirm{}: it discovers
context-specific clusters that are different across the
learned subsystems.
\cref{table:gene-sim} lists various co-clustering probabilities
between G235131 (\cref{table:gene}) and other genes, which show that a
pair of genes that are similar in the \domain{Localization} context
need not be similar in the \domain{Class} context.
Further, even though G235131 belongs to an unknown \domain{Class}, the
\hirm{} is still able to compute its co-clustering probabilities
within this context by using observations of its other properties
(i.e., relation values) that are inferred to be predictive of the
missing value.

\begin{table}[t]
\centering
\captionsetup{skip=0pt}
\caption{Example gene posterior co-clustering probabilities in each
of the two contexts shown in \cref{fig:genes-cluster-gene}.}
\label{table:gene-sim}
\footnotesize
\begin{adjustbox}{max width = \linewidth}
\begin{tabular}{|l|l|c|c|l|}
\hline
\multirow{2}{*}{\domain{Gene} 1}
  & \multirow{2}{*}{\domain{Gene} 2}
  & \multicolumn{2}{>{\centering}p{.25\textwidth}|}{Co-clustering probability \newline within subsystem containing}
  & \multirow{2}{*}{Pattern} \\ \cline{3-4}
~ & ~  & \domain{Localization} & \domain{Class} & ~ \\ \hline\hline
G235131 & G235278 &  0.98 & 0.87 & LL \\
G235131 & G239017 &  0.52 & 0.47 & MM \\
G235131 & G236063 &  0.03 & 0.13 & UU \\
G235131 & G235388 &  0.83 & 0.27 & LU \\
G235131 & G240065 &  0.03 & 0.68 & UL \\ \hline
\multicolumn{5}{@{}l}{
U = Unlikely 0--0.33;
M = Medium 0.33--0.66;
L = Likely 0.66--1
}
\end{tabular}
\end{adjustbox}
\vspace{-.5cm}
\end{table}

We next computed posterior co-clustering probabilities for domains that
represent gene properties.
In \cref{fig:genes-cluster-localization}, the \hirm{} infers a
likely cluster of \domain{Localization} entities that includes
cell wall, extracellular, integral membrane, and lipid particles,
whereas cytoplasm and nucleus are inferred as probable singletons.
\Cref{fig:genes-cluster-class} shows co-clustering probabilities
for \domain{Class}, which reflect a probable cluster (cyclins,
tublins, adaptins, \dots) embedded within a larger more noisy cluster,
as well as singletons such as transcription factor and polymerases.
These heatmaps show quantitative estimates of
posterior uncertainty in the partition structures detected by the \hirm{},
which cannot be captured using inference approaches such as approximate
maximum likelihood or maximum a posteriori estimation and
highlight a key benefit of using fully Bayesian sampling approaches
(\cref{sec:hirm-inference}) for probabilistic structure learning in
complex  domains.

\section{Related Work}
\label{sec:related-work}

Several variations of the standard IRM have been introduced in the literature
on nonparametric relational Bayesian
models~\citep{ishiguro2012,ohama2013,jonas2015,briercliffe2016}.
Our method is distinguished by being the first hierarchical extension
that uses a nonparametric structure learning prior over the relations
themselves to improve modeling capacity and address shortcomings of
the IRM identified in \cref{sec:limitations}, which include combinatorial
over-clustering and failing to detect relationships between dependent
but non-identically distributed relations.
These limitations have not been addressed by previous variations of
the IRM.
A key advantage of our hierarchical approach is that it can be
composed with several IRM variants that address other shortcomings of
the standard IRM, including
\begin{enumerate*}[label=(\roman*)]
\item the subset IRM~\citep{ishiguro2012}, which detects and filters
out irrelevant observations in the case of extreme sparsity;
and
\item the logistic regression IRM~\citep{jonas2015}, which improves
predictive accuracy for semi-supervised tasks that specify one or more
target variables as well as exogenous (non-probabilistic) predictor
variables.
\end{enumerate*}

Other approaches to relational modeling include relational extensions of
Bayesian networks~\citep{heckermen2004,koller1997,friedman1999} and Markov random fields~\citep{taskar2002,richardson2006}.
While these approaches are typically more expressive than the models we consider
here, they inherit traditional challenges of structure learning and
model selection for directed models~\citep{daly2011} (e.g., there is a
super-exponential number of graphs to consider~\citep{robinson1977});
and can require tuning evaluation measures, clause construction
operators, or search strategies~\citep{kok2005} for undirected models.
We instead build on Bayesian nonparametric relational
models~\citep{fan2020} that
\begin{enumerate*}[label=(\roman*)]
\item use latent variables to provide a layer of indirection
and simplify the learning problem as compared to searching over
arbitrary graphical structures; and
\item can be learned using principled algorithms for Bayesian inference.
\end{enumerate*}

Deep generative models have also been developed for relational
data~\citep{kipf2016,mehta2019,fan2019,qu2019}.
These methods either typically assume that there is one binary adjacency
matrix being modeled (i.e., a random graph relation) or work
in a semi-supervised setting of predicting labels.
In contrast, we aim to discover generative models for datasets with
richer relational schemas than a single binary matrix (e.g.,
\cref{fig:systems}) and operate in a fully unsupervised setting
without assuming beforehand that there are specific labels to predict.
This approach allows us in
\cref{sec:evaluations-object-attribute} to make predictions using
inferred joint probabilities for up to 1556 variables, and in
\cref{sec:evaluations-political,sec:evaluations-genomics} to automatically
model sparse and noisy systems with multiple entities, attributes, and interactions.


Using the Chinese restaurant process as a structure learning prior (\cref{eq:hirm-crp-outer})
has been considered in other settings, including
non-relational tabular data~\citep{mansinghka2016},
multivariate time series~\citep{saad2018},
topic modeling~\citep{blei2010},
and computer vision~\citep{salakhutdinov2013}, among others.
The same insight of using an outer CRP to partition relations (used in
this work to extend the IRM) can also be applied to other models that
handle relational systems with multiple relations, such as the
Mondrian process~\citep{roy2008}.
More broadly, it would be particularly fruitful to investigate a
representation theorem for the ergodic distributions of a relational
system modeled by an \hirm{} within the framework of exchangeable
random structures from~\citet{orbanz2013}.

In addition to the IRM, several other Bayesian nonparametric models are
special cases of the \hirm{}, including the
infinite hidden relational model~\citep{xu2006},
infinite mixture model~\citep{rasmussen1999},
Dirichlet process mixture model~\citep{lo1984},
and Cross-Categorization~\citep{mansinghka2016}.
By generalizing the likelihood term in \cref{eq:hirm-relation} to
include regression on relation values that are endogenous to the
system, the \hirm{} could be further extended to express a relational
variant of Dirichlet process mixtures of generalized linear
models~\citep{hannah2011}.

Finally, as a domain-general model for relational data, the \hirm{}
can be used to extend previous methods for automatic Bayesian modeling
of non-relational tabular data that synthesize probabilistic
programs in domain-specific languages~\citep{saad2019}.
Expressing the \hirm{} in probabilistic programming languages would
simplify several end-user workflows for data analysis tasks such as
imputation, outlier detection, dependence detection, and
search~\citep{saad2016,saad2017,saad2017search}, as well as enable
fast exact inference~\citep{saad2021} for the broad range of
probabilistic queries that the \hirm{} can handle.

\section{Conclusion}
\label{sec:conclusion}

This paper has presented the hierarchical infinite relational model
(\hirm{}), a new method for discovering probabilistic structure in
relational data.
A key insight in our approach is to use a nonparametric prior that
divides a system of relations into independent subsystems, each to
be learned using a separate infinite relational model.
This Bayesian nonparametric approach to structure learning generalizes
the standard infinite relational model~\citep{kemp2006} and addresses
several limitations in its inductive biases.

While methods based on the IRM, such as the \hirm{}, specify
relatively simple probabilistic theories for relational systems as
compared to other approaches that specify more complex
theories~\citep{muggleton1994,getoor2007}, our evaluations illustrate
the efficacy of our approach on density estimation tasks and show that
it can discover meaningful structure in real-world politics and
genomics datasets.
The results also underscore the benefit of principled and fully
Bayesian structure learning for inferring probable independences,
which can improve scalability, interpretability, uncertainty
characterization, and model fit.

\bibliography{paper}

\end{document}